\documentclass[journal]{IEEEtran}
\usepackage{amsmath,amsfonts}
\usepackage{algorithmic}
\usepackage{array}
\usepackage[caption=false,font=normalsize,labelfont=sf,textfont=sf]{subfig}
\usepackage{textcomp}
\usepackage{stfloats}
\usepackage{url}
\usepackage{verbatim}
\usepackage{graphicx}
\def\BibTeX{{\rm B\kern-.05em{\sc i\kern-.025em b}\kern-.08em
    T\kern-.1667em\lower.7ex\hbox{E}\kern-.125emX}}
\usepackage{balance}
\usepackage{amssymb}
\usepackage{booktabs}
\usepackage{multirow}
\usepackage{cleveref}
\usepackage{makecell}
\usepackage[table]{xcolor}
\usepackage[T1]{fontenc}
\usepackage{float}

\begin{document}

\title{Long-tailed Distribution-aware Router For Mixture-of-Experts in Large Vision-Language Models}

\author{
    Chaoxiang~Cai, Longrong~Yang, Minghe~Weng, Xuewei~Li, Zequn~Qin$^{*}$, Xi~Li$^{*}$

    \IEEEcompsocitemizethanks{\IEEEcompsocthanksitem
        Chaoxiang~Cai and Zequn~Qin are with the School of Software Technology, Zhejiang University, Ningbo 315100, China.
        Longrong~Yang, Minghe~Weng and Xi~Li are with the College of Computer Science and Technology, Zhejiang University, Hangzhou 310027, China.
        Xuewei~Li is with the School of Electronic and Information, Shanghai Dianji University, Shanghai 201306, China. 
        E-mail: \{cxcai, longrongyang, wengminghe, xilizju\}@zju.edu.cn, zequnqin@gmail.com, xueweili@sdju.edu.cn.
        \protect
    }
    \thanks{(Corresponding authors: Zequn~Qin and Xi~Li)}
}

\markboth{}%
{Shell \MakeLowercase{\textit{et al.}}: Long-tailed Distribution-aware Router For Mixture-of-Experts in Vision-Language Models}

\maketitle

\begin{abstract}
The mixture-of-experts (MoE) architecture, which replaces dense networks with sparse ones, has attracted significant attention in large vision-language models (LVLMs) for achieving comparable performance while activating far fewer parameters.
Existing MoE architectures for LVLMs primarily focus on token-to-expert routing (TER), encouraging different experts to specialize in processing specific tokens.
However, these methods typically rely on the load balancing mechanism, neglecting the inherent distributional differences between vision and language modalities.
To address this limitation, we propose the \textbf{L}ong-\textbf{T}ailed \textbf{D}istribution-aware \textbf{R}outer (LTDR) for vision-language TER, which tackles two key challenges:
(1)~Modality-specific distribution-aware routing.
We observe that language TER generally follows a relatively uniform distribution, whereas vision TER exhibits a long-tailed distribution.
This modality discrepancy motivates the design of specialized routing strategies for each modality.
(2)~Vision-specific dynamic expert activation.
Recognizing the importance of high-information vision tail tokens, we introduce a data-augmentation-inspired strategy that increases the number of activated experts, ensuring sufficient learning for these rare but informative tokens.
On vision-language and vision benchmarks, our approach achieves consistent improvements, boosting performance by 1.2\% / 2.1\% on vision-language and 1.6\% on vision benchmarks.
\end{abstract}

\begin{IEEEkeywords}
Long-tailed distribution, Modality-specific routing, Large Vision-Language Models, Mixture-of-Experts.
\end{IEEEkeywords}

\section{Introduction}
\label{main_introduction}

Recent advances in large vision-language models (LVLMs)~\cite{achiam2023gpt4, meta2024llama3}, which bridge vision and language, have demonstrated impressive instruction-following and generalization capabilities.
However, real-world necessitate models that can handle diverse tasks.
Traditional approaches that train separate models for each task incur significant redundancy and resource consumption.
Despite efforts to expand datasets or models~\cite{chen2023internvl, bai2023qwenvl, lu2024deepseekvl}, the demand for substantial resources remains.
The mixture-of-experts (MoE)~\cite{jacobs1991moe} architecture enables scalable parameter growth without a proportional increase in inference costs.
Its effectiveness in model scaling has been demonstrated in various recent works~\cite{dou2023loramoe, gou2023mocle, bai2023qwen, chen2024llavamole, dai2024deepseekmoe}. 
MoE-LLaVA~\cite{lin2024moellava} achieves performance comparable to LLaVA-7B and LLaVA-13B~\cite{liu2023llava} while activating only 3B parameters.

\begin{figure*}[t]
    \centering
    \includegraphics[width=1.0\linewidth]{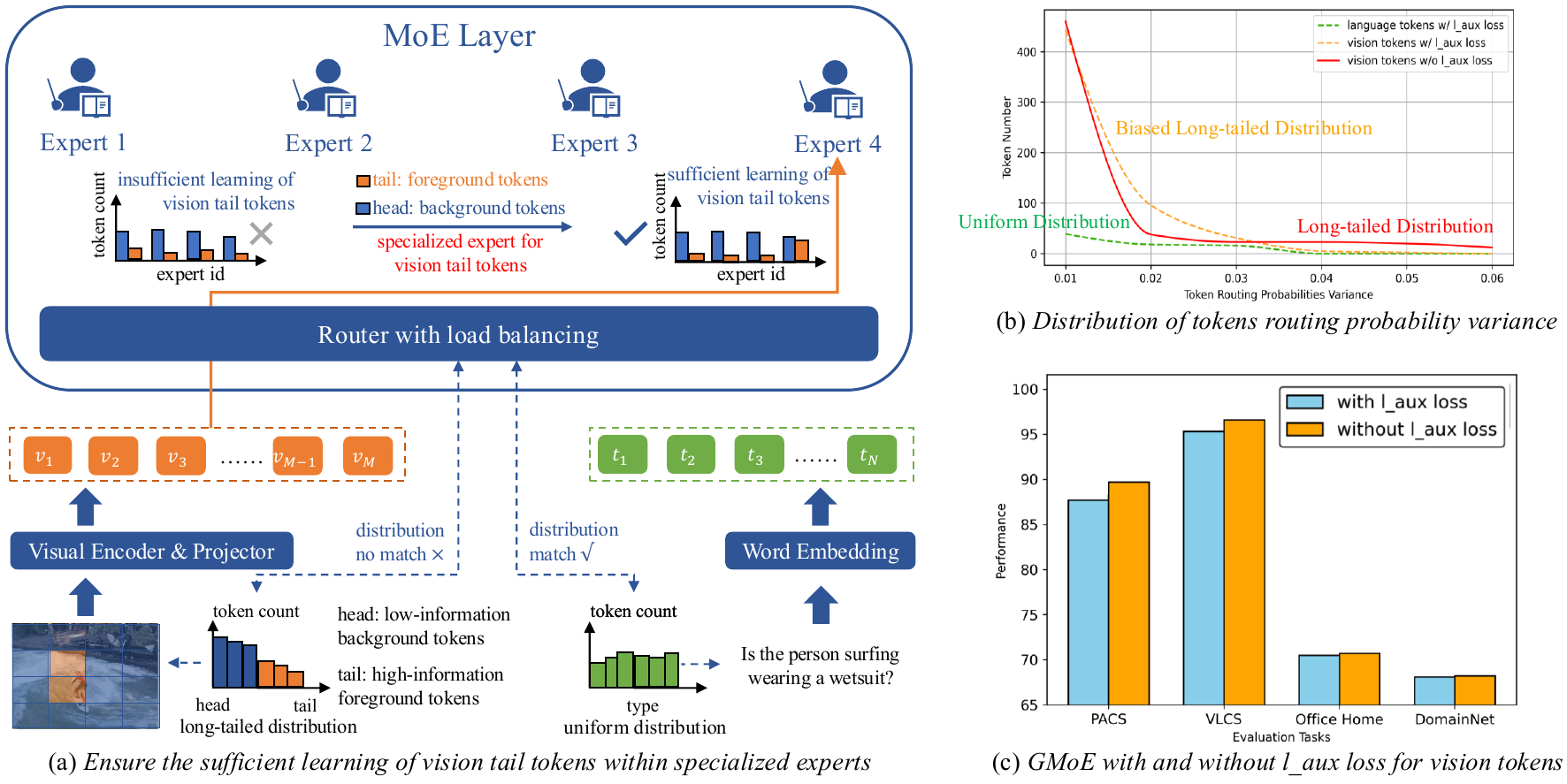}
    \caption{(a)~Our goal is to ensure that vision tail tokens are sufficiently learned within specialized experts. (b)~Distribution of TER probability variance. \textcolor[HTML]{35AE04}{Language TER with load balancing} is uniform, \textcolor{red}{vision TER without load balancing} exhibits a long-tailed distribution and \textcolor[HTML]{FEA22D}{\text{vision TER with load balancing}} shows a biased long-tailed characteristic. (c)~GMoE with and without load balancing. Removing load balancing from vision tokens improves performance.}
    \label{fig_intro}
\end{figure*}

The core of MoE lies in its token-to-expert routing (TER).
While most implementations~\cite{shazeer2017moelayer} utilize trainable routers to predict routing probabilities, they typically enforce load balancing constraints on TER to prevent expert overload or underload, thereby promoting a uniform distribution of tokens across experts.
However, this uniform-load strategy is sub-optimal for multi-modal tasks, where vision tokens follow a long-tailed distribution~\cite{he2016deep, krizhevsky2017imagenet} in contrast to the more uniform distribution of language tokens~\cite{vaswani2017transformer, devlin2019bert}. 
Indiscriminate load balancing hinders the effective learning of vision tokens.
As illustrated in Fig.~\ref{fig_intro} (a), vision tokens comprise a majority of low-information background tokens (the head) and a minority of high-information foreground tokens (the tail).
Enforcing load balancing scatters these critical, but sparse, foreground tokens across different experts, which impedes the consolidation of similar information and prevents the selection of specialized experts for these important tokens.

We perform an analysis of this question.
Fig.~\ref{fig_intro} (b) plots the TER probability variance (x-axis) against the token count (y-axis).
The distribution of vision token counts across probability variances is long-tailed.
The majority of tokens, which possess low probability variances, typically correspond to low-information background tokens, as the router struggles to assign specialized experts for them.
In contrast, high-information foreground tokens generally exhibit high variances.
As shown in Fig.~\ref{fig_intro} (b), the application of load balancing reduces the count of tokens with high variances, suggesting that the load balancing hinders the expert specialization for foreground tokens.
While language token counts are uniformly distributed across probability variances, indicating that language tokens are compatible with load balancing.
Fig.~\ref{fig_intro} (c) shows that removing load balancing for vision tokens leads to improved performance.
Based on these, our goal is to ensure that sparse yet high-information vision tail tokens are sufficiently learned within specialized experts as shown in Fig.~\ref{fig_intro} (a).

To ensure sufficient learning of vision tail tokens, we propose the \textbf{L}ong-\textbf{T}ailed \textbf{D}istribution-aware \textbf{R}outer (LTDR).
Our approach solves two key challenges:
(1)~\emph{Modality-specific distribution-aware routing.}
Increasing the variance of routing probabilities allows vision tokens, especially vision tail tokens, to be learned within specialized experts.
Consequently, we remove load balancing for vision TER to align with its long-tailed distribution, thus improving the variance of routing probabilities.
For language TER, we retain load balancing as it aligns well with its uniform distribution.
(2)~\emph{Vision-specific dynamic expert activation.}
Recognizing the importance of high-information vision tail tokens, we employ a data-augmentation strategy to increase the number of activated experts for them, thereby improving fault tolerance during  specialized expert selection.
Our method achieves improvements of 1.2\% / 2.1\% on vision-language benchmarks and 1.6\% on vision benchmarks.

Our contributions can be summarized as follows:

\begin{itemize}
    \item~We identify the detrimental effect of traditional load balancing on vision tokens. This effect stems from a distribution mismatch, where the uniform routing preference of load balancing leads to insufficient learning of important, yet long-tailed, vision tail tokens.
    
    \item~We propose the long-tailed distribution-aware router (LTDR), which contains two key modules: (1)~Modality-specific distribution-aware router for vision-language tasks with diverse distributions; and (2)~Vision-specific dynamic expert activation for vision tail tokens that  necessitate sufficient representation learning.
    
    \item~Extensive experiments demonstrate the effectiveness of our approach, achieving average improvements of 1.2\% / 2.1\% on vision-language and 1.6\% on vision benchmarks.
\end{itemize}

\section{Related Works}
\label{main_related_works}

\subsection{Mixture-of-Experts in Large Vision-Language Models}

The success of large language models (LLMs) has driven rapid progress in LVLMs~\cite{lin2024moellava}.
LVLMs~\cite{achiam2023gpt4, liu2023llava} encode visual inputs into LLM-compatible representations, enabling effective vision-language integration.
Current efforts focus on dataset scaling~\cite{zhang2023llavar} and parameter-efficient adaptation~\cite{Adapter, Prompt, LoRA, ye2023mplug} to improve efficiency.
However, LVLMs still suffer from limited generalization and high inference costs, motivating the adoption of MoE architectures.


The effectiveness of MoE models largely depends on their token-to-expert routing mechanisms.
Traditional MoE employs a trainable linear layer to predict routing probabilities~\cite{lepikhin2020gshard}, often with load-balancing constraints to ensure uniform expert utilization.
Building on this, various advanced routing strategies have been proposed.
Task-based routing~\cite{gururangan2021demix, jain2024damex, zhou2024exploring} assigns tokens to predefined experts, while cluster-based routing~\cite{dou2023loramoe, gou2023mocle, 10528872} groups feature-similar tokens to enhance specialization.
Dynamic routing~\cite{huang2024harder, guo2024dynamic} further adapts the number of activated experts, reducing reliance on fixed hyper-parameters.
Recent work also explores richer routing dynamics, including modeling task relationships~\cite{ma2018modeling}, mitigating gradient conflicts~\cite{yang2024stgc}, and sequential routing~\cite{zhong2024lory}, highlighting the critical role of routing in improving MoE efficiency.


Nevertheless, existing methods largely overlook routing under long-tailed distributions, which are common in real-world scenarios.
Prior works~\cite{wang2020long, jin2023shike} mainly address long-tailed imbalance at the sample level and rely on conventional routing strategies, without introducing dedicated routing designs.
In contrast, our LTDR targets long-tailed distributions at the token level and further enables modality-specific routing, providing a more comprehensive and effective solution.

\subsection{Long-Tailed Distribution}

Real-world data typically exhibits a long-tailed distribution, causing models to perform well on head classes while overfitting tail classes~\cite{liu2019large, cui2019class,kang2020exploring, menon2020long, song2025head, jin2025causal, 10891441}.
Long-tailed visual classification addresses this challenge under severe class imbalance.
Traditional methods can be grouped into re-sampling, re-weighting, and ensemble methods.

(1) Re-sampling methods~\cite{han2005borderline, buda2018systematic} balance data via over-sampling tail classes or under-sampling head classes.
Approaches such as BBN~\cite{zhou2020bbn} adopt dual-branch designs to jointly learn original and re-balanced distributions, while other methods use collaborative branches with loss adjustments to improve both head and tail performance.
(2) Re-weighting methods~\cite{lin2017focal, wang2017learning} assign larger weights to tail samples and down-weight head classes, typically based on inverse class frequency.
Advanced designs, such as LDAM~\cite{cao2019learning} and EQL~\cite{tan2020equalization}, further improve performance by introducing margin-based objectives or selectively ignoring gradients for rare categories.
(3) Ensemble methods~\cite{ren2020balanced, hong2021disentangling} leverage multiple experts to address long-tailed learning.
Methods like RIDE~\cite{wang2020long} reduce variance through independent expert training and dynamic routing, while other approaches employ multi-stage or multi-expert fusion strategies to balance head and tail performance and identify hard samples more effectively.

\section{Methodology}
\label{main_method}

\subsection{Preliminaries}

\textbf{Large Vision-Language Models.}
LVLMs integrate the capabilities of LLMs with visual processing technologies, enabling vision-language understanding and generation.
As shown in Fig.~\ref{fig_intro} (a), the text input is first transformed through a word embedding, which projects the text $\mathbf{t}$ into a continuous vector space, resulting in the language token sequence $\mathcal{T} = [t_{1}, t_{2}, \cdots, t_{N}] \in \mathbb{R}^{N\times D}$.
$N$ means the sequence length of language tokens, and $D$ denotes the hidden layer size of the LLM.

Similarly, the RGB image input $\mathbf{v} \in \mathbb{R}^{H \times W \times 3}$, where $H$, $W$, and $3$ denote the height, width, and channels of the image at its original resolution, the visual encoder processes the image to extract a sequence of vision tokens $\mathcal{Z} = [z_{1}, z_{2}, \cdots, z_{M}] \in \mathbb{R}^{M \times C}$.
$M$ is the sequence length of vision tokens, and $C$ is the hidden layer size of the visual encoder.
To align vision tokens with language tokens in the same vector space, a visual projection layer is employed to map $\mathcal{Z} \in \mathbb{R}^{M \times C}$ to $\mathcal{V} = [v_{1}, v_{2}, \cdots, v_{M}] \in \mathbb{R}^{M \times D}$, where $D$ matches the hidden layer size of the LLM.
This alignment ensures that both vision and language can be processed jointly by the subsequent layers of the model.

Subsequently, vision and language tokens are concatenated and fed into the LLM.
The LLM consists of layers of multi-head self-attention (MSA) and feed-forward network (FFN), with layer normalization (LN) and residual connection applied to stabilize training and enhance performance.
As shown in Eq.~\ref{eq1} $\sim$ Eq.~\ref{eq4}, where $L$ is the layer number of LLM, the LVLM achieve a deep understanding of the relationships between vision and language, enabling effective performance on vision-language tasks.
\begin{equation}
    \mathbf{x}_0 = [v_{1}, v_{2}, \cdots, v_{M}, \cdots, t_{1}, t_{2}, \cdots, t_{N}]
    \label{eq1}
\end{equation}
\begin{equation}
    \mathbf{x}_{\ell}^{\prime} =\mathrm{MSA}(\mathrm{LN}(\mathbf{x}_{\ell-1}))+\mathbf{x}_{\ell-1}, \ell \in \{1, \ldots, L\}
    \label{eq2}
\end{equation}
\begin{equation}
    \mathbf{x}_{\ell} =\mathrm{FFN}(\mathrm{LN}(\mathbf{x^{\prime}}_{\ell}))+\mathbf{x^{\prime}}_{\ell}, \ell \in \{1, \ldots, L\}
    \label{eq3}
\end{equation}
\begin{equation}
    \mathcal{Y}=\mathrm{LN}(\mathbf{x}_L)
    \label{eq4}
\end{equation}
The output of the LVLM is optimized through a generative loss in an auto-regressive manner.
Given an image and its corresponding instruction text, the LVLM aims to generate the output text sequence $\mathcal{Y} = [y_{1}, y_{2}, \cdots, y_{O}] \in \mathbb{R}^{O\times D}$ by progressively prediction, where $O$ is the sequence length of the text output.
The loss function is defined in Eq.~\ref{eq5}, $\mathcal{Y}_{<i}$ indicates the output sequence before token $y_i$, $\theta$ denotes the trainable parameters of the model.
We only calculate the loss for the generated text.
\begin{equation}
    \mathcal{L}_{\text{regressive}}=-\sum_{i=1}^O \text{log} \ p \left( y_{i} \mid \mathcal{V}, \mathcal{T}, \mathcal{Y}_{<i}, \theta \right)
    \label{eq5}
\end{equation}

\textbf{Mixture-of-Experts.}
We replace FFN layers as MoE layers, following MoE-LLaVA~\cite{lin2024moellava}.
A MoE layer typically contains multiple FFNs, denoted as an experts ensemble $\mathcal{E} = [e_{1}, e_{2}, \cdots, e_{K}]$, $K$ is the number of total experts.
The router implements a linear layer to predict the routing probability of assigning tokens to experts.
As shown in Eq.~\ref{eq6}, the router produces weight logits \(f(\mathbf{x}) = \mathbf{W} \cdot \mathbf{x} \), which are normalized by the softmax function.
The matrix \(\mathbf{W} \in \mathbb{R}^{D \times K}\) denotes the lightweight trainable parameters for routing, and \(\mathcal{P}(\mathbf{x})_i \) is the routing score of the input \(\mathbf{x}\) for the \(i\)-th expert.
The final output in Eq.~\ref{eq7} is computed as a weighted sum of the outputs from the Top-\(k\) experts with the highest softmax probabilities.
\(\mathcal{E}(\mathbf{x})_i \) is the output of the \(i\)-th expert, and the weight for each expert is determined by its routing score.
\begin{equation}
    \mathcal{P}(\mathbf{x})_i = \frac{e^{f(\mathbf{x})_i}}{\sum_{j=1}^K e^{f(\mathbf{x})_j}}
    \label{eq6}
\end{equation}
\begin{equation}
    \mathrm{MoE}(\mathbf{x}) = \sum_{i=1}^k \mathcal{P}(\mathbf{x})_i \cdot \mathcal{E}(\mathbf{x})_i
    \label{eq7}
\end{equation}
Due to the presence of multiple experts, it is necessary to impose the expert load balancing constraint on MoE layers.
Traditional methods~\cite{lin2024moellava, yang2024stgc} incorporate differentiable load balancing loss~\cite{fedus2022switch} into each MoE layer to encourage experts to handle tokens in a balanced manner.
As shown in Eq.~\ref{eq8}, \(\mathcal{F}_i \) is the fraction of tokens processed by expert \(\mathcal{E}_i \), and \(\mathcal{G}_i \) is the average routing probability of expert \(\mathcal{E}_i \).
\begin{equation}
    \mathcal{L}_{\text{balancing}} = K \cdot \sum_{i=1}^K \mathcal{F}_{i} \cdot \mathcal{G}_{i}
    \label{eq8}
\end{equation}

\begin{figure*}[t]
    \centering
    \includegraphics[width=1.0\linewidth]{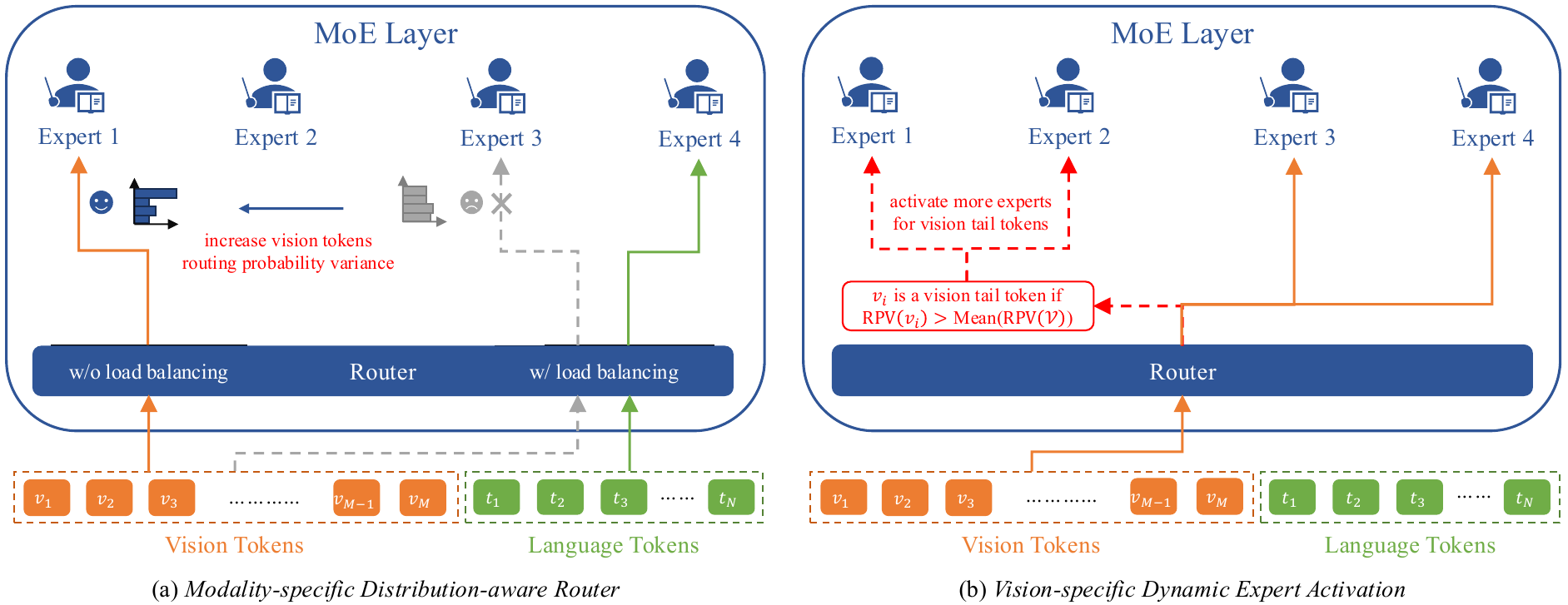}
    \caption{(a)~Modality-specific distribution-aware router allows vision and language to be routed with different expert load to adapt to their respective modality distributions. (b)~Vision-specific dynamic expert activation enables a data-augmentation strategy to make experts process important vision tail tokens sufficiently.}
    \label{fig_method}
\end{figure*}

\textbf{Our Method LTDR.}
Since language tokens follow a uniform distribution, while vision tokens exhibit a long-tailed distribution, we focus on optimizing vision-language TER to make experts handle different distributional modality tokens effectively.
We find that the load balancing mechanism leads to the scattered vision tail tokens in experts, impeding the learning of specialized experts.
Therefore, as illustrated in Fig.~\ref{fig_method}, our method consists of two modules:
(1)~Modality-specific Distribution-aware Router (MsDaR).
We retain load balancing for language TER as it aligns with the uniform distribution of language tokens, while abandon load balancing for vision TER to adaptively align with the long-tailed distribution of visual tokens.
Without load balancing, vision tokens, especially vision tail tokens, exhibit higher routing probability variance, enabling expert specialization.
(2)~Vision-specific Dynamic Expert Activation (VsDEA).
Given the high importance of vision tail tokens, we define the head and tail tokens of vision, and increase the number of activated experts for vision tail tokens, achieving a data-augmentation strategy to improve fault tolerance and learning effectiveness.

\subsection{Modality-specific Distribution-aware Router}

Existing MoE architectures on modality differences fall into modality-aware~\cite{nguyen2024expert, chen2024eve, lin2024moma} and distribution-aware~\cite{wang2020long, jin2023shike}.
~\cite{nguyen2024expert, lin2024moma} center on modalities by using a hierarchical MoE and modality-specific expert groups, while~\cite{wang2020long, jin2023shike} focus on long-tailed distribution by enhancing expert diversity and reducing dynamic routing.
These methods are constrained by load balancing (Eq.~\ref{eq8}), without considering modality token distribution differences.

Recent works~\cite{he2016deep, krizhevsky2017imagenet, vaswani2017transformer, devlin2019bert} have shown that language follows a uniform distribution, whereas vision exhibits a long-tailed distribution.
This divergence stems from the characteristics of vision, which contains a few foreground patches and a large number of background patches.
~\cite{radford2021learning, kim2021vilt} also emphasize that the structural and semantic differences between vision and language necessitate specific processing.
To this end, we optimize the TER distribution (Eq.~\ref{eq6}) for vision and language.

Load balancing conflicts with the long-tailed vision distribution, an intuitive way is to release vision tokens from load balancing to increase their routing probability variance (RPV).
As shown in Eq.~\ref{eq9}, for a vision token \(v_i \in \mathcal{V} \), its routing probabilities \(\mathcal{P}(v_i) \in \mathbb{R}^{K} \), and \(\mathrm{RPV}(v_i) \) is its variance of \(\mathcal{P}(v_i) \).
The number of vision head tokens is sufficient to generalize across various experts, resulting in uniform routing probability (low RPV).
If the limited number of vision tail tokens are distributed uniformly across experts, it would lead to poor generalization.
Therefore, a modality-specific distribution-aware router is essential to benefit the sufficient learning of experts for vision tokens.
We modify the traditional \(\mathcal{L}_{\text{balancing}} \) in Eq.~\ref{eq10}, where \(\mathcal{T} \) means the language token sequence.
\begin{equation}
    \mathrm{RPV}(v_i) = \mathrm{Variance}(\mathcal{P}(v_i))
    \label{eq9}
\end{equation}
\begin{equation}
    \mathcal{L}_{\text{balancing}} = \sum\limits_{i=1}^K \mathcal{F}_{i}(\mathcal{T}) \cdot \mathcal{G}_{i}(\mathcal{T})
    \label{eq10}
\end{equation}

In this architecture, language tokens maintain their original load balancing behavior, while vision tokens are released from the constraint of load balancing.
This enables vision tokens, especially vision tail tokens, which represent information-rich content, to undergo specialized expert processing.
As shown in Fig.~\ref{fig_method} (a), by enhancing the RPV of vision tokens, these tokens can be allocated to specialized experts instead of being uniformly distributed, facilitating more specialized and efficient learning.

\subsection{Vision-specific Dynamic Expert Activation}

Given the high importance of vision tail tokens, we enhance their processing for sufficient expert learning.
However, the complexity and training instability introduced by long-tailed tokens necessitate a simple yet effective solution.
Since the input and output token sequence lengths fix before and after each FFN layer, traditional sampling methods are impractical for adjusting head or tail token counts.
We thus employ a data-augmentation strategy to enhance the expert learning and processing for vision tail tokens.
As illustrated in Fig.~\ref{fig_method} (b), our innovation lies in defining and identifying vision tail tokens and activating more experts to learn and process vision tail tokens. First, we define whether a vision token \(v_i \in \mathcal{V} \) is a vision head token or vision tail token by the following Eq.~\ref{eq11}.
\begin{equation}
    \text{Type}(v_i) = \text{tail}, \text{if } \mathrm{RPV}(v_i) > \mathrm{Mean}(\mathrm{RPV}(\mathcal{V}))
    \label{eq11}
\end{equation}
\(\mathrm{Mean}(\mathrm{RPV}(\mathcal{V})) \) is the mean value of \(\mathrm{RPV}(\mathcal{V}) \in \mathbb{R}^M \).
We serve vision tokens with larger \(\mathrm{RPV} \) than the \(\mathrm{Mean}(\mathrm{RPV}(\mathcal{V})) \) of total vision tokens as vision tail tokens, and the remainder as heads.
Vision tail tokens generally exhibit higher \(\mathrm{RPV} \) than head tokens, as they are processed by specialized experts.
We use mean \(\mathrm{RPV} \) as the dynamic threshold for vision tail token recognition, as it filters out most of vision head tokens.
Subsequently, we route identified vision tail tokens to more experts than originally assigned, resulting in a modified version of Eq.~\ref{eq7} tailored for vision tail tokens in Eq.~\ref{eq12}.

\begin{equation}
    \begin{split}
    \mathrm{MoE}(x) & =
    \begin{cases} 
    \sum_{i=1}^a \frac{e^{\mathcal{P}(x)_i}}{\sum_{j=1}^a e^{\mathcal{P}(x)_j}} \cdot \mathcal{E}(x)_i, & \text{vision tail token} \\
    \sum_{i=1}^k \frac{e^{\mathcal{P}(x)_i}}{\sum_{j=1}^k e^{\mathcal{P}(x)_j}} \cdot \mathcal{E}(x)_i, & \text{others}
    \end{cases}
    \end{split}
    \label{eq12}
\end{equation}
The weights of the selected experts will be renormalized.
Vision tail tokens can be handled by more experts (\(k < a \leq K \)).
Reducing the possibility of expert incorrect routing.

\begin{table*}[ht]
    \caption{Comparisons of different LVLMs and our method. MoE-LLaVA uses 4Top2 while our method with the VsDEA module uses Top4. We calculate the average performance ``Avg'' across all datasets except for MME.}
    \label{tab_vl1}
    \centering
    \resizebox{\textwidth}{!}{
        \begin{tabular}{l|c|ccccccc|c}
            \toprule
            \textbf{Method} & \textbf{LLM} & \textbf{GQA} & \textbf{ScienceQA-IMG} & \textbf{TextVQA} & \textbf{POPE} & \textbf{MME} & \textbf{MMBench} & \textbf{MM-Vet} & \textbf{Avg} \\
            \midrule
            \multicolumn{10}{c}{\textit{Dense Model}} \\
            \midrule
            \rowcolor{gray!20} \textcolor{gray}{IDEFICS}~\cite{laurenccon2023obelics} & \textcolor{gray}{IDEFICS-65B}~\cite{laurenccon2023obelics} & \textcolor{gray}{45.2} & \textcolor{gray}{-} & \textcolor{gray}{30.9} & \textcolor{gray}{-} & \textcolor{gray}{-} & \textcolor{gray}{54.5} & \textcolor{gray}{-} & \textcolor{gray}{-} \\
            \rowcolor{gray!20} \textcolor{gray}{LLaVA-1.5}~\cite{liu2023llava} & \textcolor{gray}{LLaMA-13B}~\cite{touvron2023llama} & \textcolor{gray}{63.3} & \textcolor{gray}{71.6} & \textcolor{gray}{61.3} & \textcolor{gray}{85.9} & \textcolor{gray}{1531.3} & \textcolor{gray}{67.7} & \textcolor{gray}{35.4} & \textcolor{gray}{64.2} \\
            \rowcolor{gray!20} \textcolor{gray}{LLaVA-1.5}~\cite{liu2023llava} & \textcolor{gray}{Vicuna-7B}~\cite{chiang2023vicuna} & \textcolor{gray}{62.0} & \textcolor{gray}{66.8} & \textcolor{gray}{58.2} & \textcolor{gray}{85.9} & \textcolor{gray}{1510.7} & \textcolor{gray}{64.3} & \textcolor{gray}{30.5} & \textcolor{gray}{61.2} \\
            \rowcolor{gray!20} \textcolor{gray}{Qwen-VL}~\cite{bai2023qwenvl} & \textcolor{gray}{Qwen-7B}~\cite{bai2023qwenvl} & \textcolor{gray}{59.3} & \textcolor{gray}{67.1} & \textcolor{gray}{63.8} & \textcolor{gray}{-} & \textcolor{gray}{-} & \textcolor{gray}{38.2} & \textcolor{gray}{-} & \textcolor{gray}{-} \\
            TinyGPT-V~\cite{yuan2023tinygpt} & Phi2-2.7B~\cite{javaheripi2023phi} & 33.6 & - & - & - & - & - & - & - \\
            MobileVLM~\cite{chu2023mobilevlm} & MobileLLaMA-2.7B~\cite{chu2023mobilevlm} & 59.0 & 61.0 & 47.5 & 84.9 & 1288.9 & 59.6 & - & - \\
            LLaVA-Phi~\cite{zhu2024llava} & Phi2-2.7B~\cite{javaheripi2023phi} & - & 68.4 & 48.6 & 85.0 & 1335.1 & 59.8 & 28.9 & - \\
            \midrule
            \multicolumn{10}{c}{\textit{Sparse Model}} \\
            \midrule
            MoE-LLaVA-4Top2~\cite{lin2024moellava} & StableLM-1.6B~\cite{bellagente2024stable} & 60.3 & 62.6 & 50.1 & 85.7 & 1318.2 & 60.2 & 26.9 & 57.6 \\
            \textbf{Our Method} & \textbf{StableLM-1.6B}~\cite{bellagente2024stable} & \textbf{61.1} & \textbf{63.4} & \textbf{51.1} & \textbf{86.6} & \textbf{1363.5} & \textbf{60.6} & \textbf{29.9} & \textbf{58.8} \\
            \midrule
            MoE-LLaVA-4Top2~\cite{lin2024moellava} & Phi2-2.7B~\cite{javaheripi2023phi} & 61.4 & 68.5 & 51.4 & 86.3 & 1423.0 & 65.2 & 34.3 & 61.1 \\
            \textbf{Our Method} & \textbf{Phi2-2.7B}~\cite{javaheripi2023phi} & \textbf{62.2} & \textbf{69.3} & \textbf{52.9} & \textbf{87.5} & \textbf{1446.5} & \textbf{66.8} & \textbf{34.9} & \textbf{62.3} \\
            \bottomrule
        \end{tabular}
    }
\end{table*}

\begin{table*}[ht]
    \caption{Comparisons of our method with Molmo and GMoE. Molmo adopts a 64Top8 configuration, whereas our method, enhanced by the VsDEA module, uses a more efficient 64Top12. Similarly, GMoE operates on 4Top2/6Top2, while our approach achieves comparable results with only Top4. We calculate the average performance ``Avg'' across all datasets.}
    \label{tab_vl_v}
    \centering
    \resizebox{\textwidth}{!}{
        \begin{tabular}{l|c|cccccc|c}
            \toprule
            \multicolumn{9}{c}{\textbf{Vision-Language Evaluation}} \\
            \midrule
            \textbf{Method} & \textbf{MoE Topk} & \textbf{ChartQA} & \textbf{DocVQA} & \textbf{AI2D} & \textbf{VQA} & \textbf{AndroidControl} & \textbf{CountBenchQA} & \textbf{Avg} \\
            \midrule
            Molmo~\cite{deitke2024molmo} & Top8 & 65.7 & 79.8 & 85.2 & 82.6 & 81.8 & 74.0 & 78.1 \\
            \textbf{Our Method} & \textbf{Top12 (VsDEA) / Top8 (others)} & \textbf{68.1} & \textbf{81.3} & \textbf{87.4} & \textbf{83.3} & \textbf{83.2} & \textbf{77.6} & \textbf{80.2} \\
            \toprule
            \multicolumn{9}{c}{\textbf{Vision Evaluation}} \\
            \midrule
            \multirow{2}{*}{\textbf{Method}} & \multirow{2}{*}{\textbf{MoE Topk}} & \multicolumn{2}{c|}{\textbf{PACS}} & \multicolumn{2}{c|}{\textbf{VLCS}} & \multicolumn{2}{c|}{\textbf{Office-Home}} & \multirow{2}{*}{\textbf{Avg}} \\
            \cmidrule(lr){3-4} \cmidrule(lr){5-6} \cmidrule(lr){7-8}
            & & 4Top2 & \multicolumn{1}{c|}{6Top2} & 4Top2 & \multicolumn{1}{c|}{6Top2} & 4Top2 & \multicolumn{1}{c|}{6Top2} & \\
            \midrule
            GMoE~\cite{li2022sparse} & Top2 & 88.9 & \multicolumn{1}{c|}{87.6} & 92.5 & \multicolumn{1}{c|}{96.5} & 70.5 & \multicolumn{1}{c|}{71.0} & 84.5 \\
            \textbf{Our Method} & \textbf{Top4 (VsDEA) / Top2 (others)} & \textbf{90.1} & \multicolumn{1}{c|}{\textbf{92.1}} & \textbf{95.7} & \multicolumn{1}{c|}{\textbf{96.9}} & \textbf{70.6} & \multicolumn{1}{c|}{\textbf{71.4}} & \textbf{86.1} \\
            \bottomrule
        \end{tabular}
    }
\end{table*}

\section{Experiments}
\label{main_experiments}

\subsection{Experimental Setups}

\textbf{Benchmarks.}
We aim to study the vision TER within the MoE architecture.
To ensure a robust evaluation, we conduct experiments on a suite of both vision-language and vision tasks.
A detailed description of the benchmarks and tuning datasets can be found in the supplementary material.
This comprehensive evaluation framework allows for a rigorous assessment of our approach across a diverse capabilities.

\textbf{Baselines.}
We employ MoE-LLaVA~\cite{lin2024moellava} and Molmo~\cite{deitke2024molmo} for vision-language tasks, and GMoE~\cite{li2022sparse} for vision tasks.
MoE-LLaVA selects the two most relevant experts from a total of four experts, while Molmo selects 8 experts from 64 experts.
GMoE is designed for visual generalization and enhances the ability of cross-domain data generalization.

\textbf{Configurations.}
For vision-language tasks, we build upon two architectures: MoE-LLaVA (with StableLM-1.6B and Phi-2-2.7B backbones) and Molmo (with OLMoE-1B-7B backbone).
For vision tasks, we adopt the pre-trained ViT-S/16 backbone following GMoE.
Besides, our routing strategy is applied at each training batch.
Further implementation details are provided in the supplementary material.

\subsection{Comprehensive Evaluations}
\label{Comprehensive_Evaluations}

\textbf{Vision-Language.}
As shown in Tab.~\ref{tab_vl1} $\sim$ Tab.~\ref{tab_vl_v}, our method demonstrates robust image-text understanding, achieving superior performance compared to the MoE-LLaVA and Molmo.
Specifically, our method achieves average improvements of 1.2\% over StableLM-1.6B, 1.2\% over Phi2-2.7B, and 2.1\% over OLMoE-1B-7B.
The improvements across different models also highlight the scalability of our method.

\textbf{Vision.}
Tab.~\ref{tab_vl_v} assesses the generalization capability of our method on vision tasks.
Our method enhances the performance of GMoE, yielding an average improvement of 1.6\%.

\subsection{Ablation Studies}

\textbf{Accuracy.}
Tab.~\ref{tab_ablation} show ablation studies on MoE-LLaVA and Molmo.
The results confirm the importance of modality-specific distribution-aware router (MsDaR) and vision-specific dynamic expert activation (VsDEA) modules.

\begin{figure*}[ht]
    \centering
    \includegraphics[width=1.0 \linewidth]{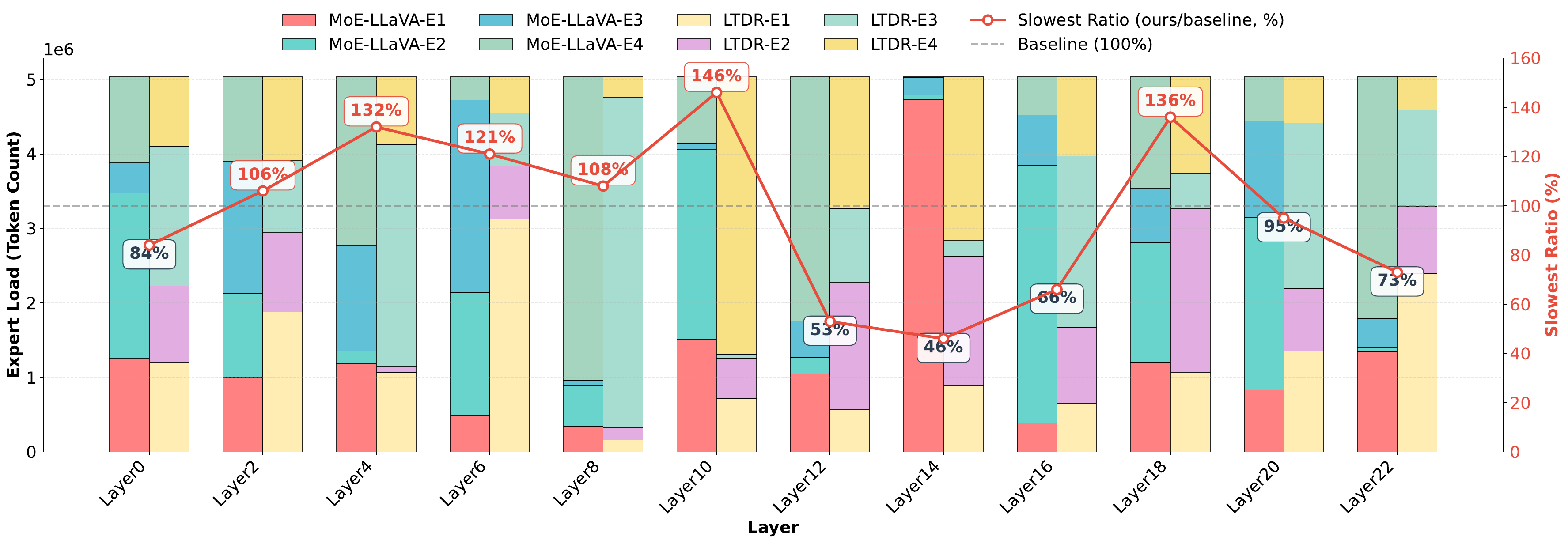}
    \caption{Expert load of MoE-LLaVA with StableLM-1.6B and LTDR on MME. LTDR does not significantly increase the load on the slowest experts.}
    \label{fig_expertloading}
\end{figure*}

\textbf{Running Time.}
As shown in Tab.~\ref{tab_runningtime}.
While our method activates more experts for vision tail tokens, its inference time does not increase significantly.
This stems from the all-to-all communication waiting principle: our method enhances expert activation without overburdening the slowest expert. Details of expert load is provided in Fig.~\ref{fig_expertloading}.

\begin{table*}[ht]
    \caption{Modular ablation studies on MoE-LLaVA and Molmo.}
    \label{tab_ablation}
    \centering
    \resizebox{\textwidth}{!}{
        \begin{tabular}{l|cccccc|c}
            \toprule
            \textbf{Method} & \textbf{GQA} & \textbf{ScienceQA-IMG} & \textbf{TextVQA} & \textbf{POPE} & \textbf{MMBench} & \textbf{MM-Vet} & \textbf{Avg} \\
            \midrule
            MoE-LLaVA-4Top2 (StableLM-1.6B) & 60.3 & 62.6 & 50.1 & 85.7 & 60.2 & 26.9 & 57.6 \\
            + MsDaR & 61.1 & 62.3 & 51.2 & 86.6 & 59.9 & 27.9 & 58.2 \\
            + MsDaR\&VsDEA (LTDR) & 61.1 & 63.4 & 51.1 & 86.6 & 60.6 & 29.9 & 58.8 \\
            \midrule
            MoE-LLaVA-4Top2 (Phi2-2.7B) & 61.4 & 68.5 & 51.4 & 86.3 & 65.2 & 34.3 & 61.1 \\
            + MsDaR & 61.8 & 68.6 & 51.8 & 86.7 & 66.2 & 34.5 & 61.6 \\
            + MsDaR\&VsDEA (LTDR) & 62.2 & 69.3 & 52.9 & 87.5 & 66.8 & 34.9 & 62.3 \\
            \toprule
            \textbf{Method} & \textbf{ChartQA} & \textbf{DocVQA} & \textbf{AI2D} & \textbf{VQA} & \textbf{AndroidControl} & \textbf{CountBenchQA} & \textbf{Avg} \\
            \midrule
            Molmo-64Top8 (OLMoE-1B-7B) & 65.7 & 79.8 & 85.2 & 82.6 & 81.8 & 74.0 & 78.1 \\
            + MsDaR & 67.8 & 80.7 & 86.4 & 83.0 & 81.7 & 76.9 & 79.4 \\
            + MsDaR\&VsDEA (LTDR) & 68.1 & 81.3 & 87.4 & 83.3 & 83.2 & 77.6 & 80.2 \\
            \bottomrule
        \end{tabular}
    }
\end{table*}

\begin{table*}[ht]
    \caption{Running time comparison on MoE-LLaVA with StableLM-1.6B and Molmo with OLMoE-1B-7B.}
    \label{tab_runningtime}
    \centering
    \resizebox{\textwidth}{!}{
        \begin{tabular}{l|c|cccccc|c}
            \toprule
            \textbf{Method} & \textbf{GPU} & \textbf{GQA} & \textbf{ScienceQA-IMG} & \textbf{TextVQA} & \textbf{POPE} & \textbf{MMBench} & \textbf{MM-Vet} & \textbf{Avg (s)} \\
            \midrule
            MoE-LLaVA-4Top2 & V100-30G & 2284 & 331 & 1277 & 1552 & 835 & 366 & 1108 \\
            Our Method & V100-30G & 2252 & 368 & 1259 & 1530 & 825 & 363 & 1100 \\
            \midrule
            MoE-LLaVA-4Top2 & A800-80G & 1771 & 285 & 1265 & 1269 & 603 & 310 & 917 \\
            Our method & A800-80G & 1698 & 301 & 1057 & 1116 & 595 & 310 & 846 \\
            \toprule
            \textbf{Method} & \textbf{GPU} & \textbf{ChartQA} & \textbf{DocVQA} & \textbf{AI2D} & \textbf{VQA} & \textbf{AndroidControl} & \textbf{CountBenchQA} & \textbf{Avg (s)} \\
            \midrule
            Molmo-64Top8 & A800-80G & 184 & 200 & 190 & 196 & 89 & 705 & 261 \\
            Our Method & A800-80G & 187 & 203 & 194 & 200 & 90 & 716 & 265 \\
            \bottomrule
        \end{tabular}
    }
\end{table*}

\begin{table*}[!t]
    \centering
    \scriptsize
    \caption{Memory used (G) and GPU Utilization (\%) comparison on MoE-LLaVA-4Top2 with StableLM-1.6B.}
    \label{tab_efficiencycomparison}
    \resizebox{\textwidth}{!}{
        \begin{tabular}{l|cc|cc|cc|cc|cc|cc|cc|cc}
            \toprule
            \multirow{2}{*}{\textbf{Model}} & \multicolumn{2}{c|}{\textbf{GQA}} & \multicolumn{2}{c|}{\textbf{ScienceQA-IMG}} & \multicolumn{2}{c|}{\textbf{TextVQA}} & \multicolumn{2}{c|}{\textbf{POPE}} & \multicolumn{2}{c|}{\textbf{MME}} & \multicolumn{2}{c|}{\textbf{MMBench}} & \multicolumn{2}{c|}{\textbf{MM-Vet}} & \multicolumn{2}{c}{\textbf{Avg}} \\
            \cmidrule(lr){2-3} \cmidrule(lr){4-5} \cmidrule(lr){6-7} \cmidrule(lr){8-9} \cmidrule(lr){10-11} \cmidrule(lr){12-13} \cmidrule(lr){14-15} \cmidrule(lr){16-17} & Memo (G) & Util (\%) & Memo (G) & Util (\%) & Memo (G) & Util (\%) & Memo (G) & Util (\%) & Memo (G) & Util (\%) & Memo (G) & Util (\%) & Memo (G) & Util (\%) & Memo (G) & Util (\%) \\
            \midrule
            Vanilla & 9.30 & 65 & 9.08 & 61 & 10.43 & 63 & 8.63 & 67 & 9.11 & 66 & 10.47 & 64 & 9.08 & 31 & 9.44 & 59.57 \\
            Our Method & 9.32 & 67 & 9.03 & 57 & 10.43 & 59 & 8.63 & 67 & 9.11 & 67 & 10.47 & 65 & 9.08 & 33 & 9.44 & 59.29 \\
            \bottomrule
        \end{tabular}
    }
\end{table*}

\textbf{Expert Load.}
Although our method activates more experts, it does not increase inference time.
We attribute this to the all-to-all communication, in which the inference speed is determined by the slowest expert.
As shown in Fig.~\ref{fig_expertloading}, our method enhances expert activation without substantially increasing the slowest expert load, thereby preserving inference efficiency.

\textbf{Memory and Utilization.}
We quantify VsDEA's Memory usage (G) and GPU Utilization (\%) using a V100-30G GPU.
The analyzed results are shown in Tab.~\ref{tab_efficiencycomparison}.
It is worth noting that our method and the baseline show almost identical metrics, which means that our method does not introduce additional computational overhead.

\begin{table}[ht]
    \caption{Comparison of routing strategies on MoE-LLaVA-4Top2 with StableLM-1.6B. The best is marked with \textbf{boldface}.}
    \label{tab_routingstrategy}
    \centering
    \resizebox{0.49 \textwidth}{!}{
        \begin{tabular}{l|cccc|c}
            \toprule
            \textbf{Method} & \textbf{GQA} & \textbf{TextVQA} & \textbf{POPE} & \textbf{MM-Vet} & \textbf{Avg} \\
            \midrule
            Vanilla & 60.3 & 50.1 & 85.7 & 26.9 & 55.7 \\
            \midrule
            + Task~\cite{gururangan2021demix, jain2024damex, zhou2024exploring} & 58.2 & 49.2 & 81.5 & 25.2 & 53.5 \\
            + Cluster~\cite{dou2023loramoe, gou2023mocle} & 57.0 & 50.3 & 86.1 & 27.3 & 55.1 \\
            + Instruct~\cite{chen2311octavius} & 58.1 & 50.0 & 85.9 & 26.6 & 55.1 \\
            + Dynamic~\cite{huang2024harder, guo2024dynamic} & 61.0 & 49.2 & 85.7 & 28.2 & 56.0 \\
            + STGC~\cite{yang2024stgc} & 60.9 & 50.7 & 85.9 & 28.2 & 56.4 \\
            + Modality~\cite{chen2024eve, lin2024moma, nguyen2024expert} & 60.4 & 49.2 & 85.9 & 28.4 & 55.9 \\
            + Distribution~\cite{jin2023shike} & 60.1 & 50.5 & 86.5 & 27.8 & 56.2 \\
            \midrule
            + LTDR (Ours) & \textbf{61.1} & \textbf{51.1} & \textbf{86.6} & \textbf{29.9} & \textbf{57.2} \\
            \bottomrule
        \end{tabular}
    }
\end{table}

\begin{table}[ht]
    \centering
    \scriptsize
    \caption{Comparison of sharpening vision token routing strategies on MoE-LLaVA-4Top2 with StableLM-1.6B.}
    \label{tab_sharpening}
    \resizebox{0.49 \textwidth}{!}{
        \begin{tabular}{l|ccccccc}
            \toprule
            \textbf{Mehtod} & \textbf{ScienceQA-IMG} & \textbf{TextVQA} & \textbf{MME} & \textbf{MM-Vet} \\
            \midrule
            Vanilla & 62.6 & 50.1 & 1318.2 & 26.9 \\
            \midrule
            + Temperature & 62.8 & 50.2 & 1302.5 & 26.8 \\
            + Entropy & 62.4 & 50.8 & 1330.7 & 27.2 \\
            + Heuristic & 62.5 & 50.3 & 1299.6 & 26.6 \\
            \midrule
            + LTDR (Ours) & 63.4 & 51.1 & 1363.5 & 29.9 \\
            \bottomrule
        \end{tabular}
    }
\end{table}

\begin{table}[ht]
    \caption{Comparison of vision token selection strategies in VsDEA on MoE-LLaVA-4Top2 with StableLM-1.6B. Including the selection of vision head and tail tokens, as well as the selection of instruction-aware vision tail tokens. For vision tail tokens, it includes cases under different fixed ratios.}
    \label{tab_tokenselection}
    \centering
    \resizebox{0.49 \textwidth}{!}{
        \begin{tabular}{l|ccc|c}
            \toprule
            \textbf{Method} & \textbf{ScienceQA-IMG} & \textbf{TextVQA} & \textbf{MM-Vet} & \textbf{Avg} \\
            \midrule
            Vanilla & 62.6 & 50.1 & 26.9 & 57.6 \\
            \midrule
            VHTs & 62.0 & 50.3 & 27.8 & 58.0 \\
            IATs & 61.0 & 50.8 & 26.9 & 57.7 \\
            \midrule
            \multicolumn{5}{c}{VTTs} \\
            \midrule
            10\% (fixed) & 62.2 & 50.9 & 25.5 & 57.9 \\
            15\% (fixed) & 62.2 & 50.8 & 25.3 & 57.9 \\
            20\% (fixed) & 62.8 & 50.1 & 25.1 & 57.4 \\
            \midrule
            Ours ($\approx$13\%, adaptive) & 63.4 & 51.1 & 29.9 & 58.8 \\
            \bottomrule
        \end{tabular}
    }
\end{table}

\begin{table}[ht]
    \caption{Comparison with the ``Share + Routing” strategy of DeepSeekMoE~\cite{dai2024deepseekmoe} on MoE-LLaVA-4Top2 with StableLM-1.6B. \(S_{s} + R_{r}\) denotes S share expert(s) of size s combined with R routing experts of size r. The best is marked with \textbf{boldface}.}
    \label{tab_deepseek}
    \centering
    \resizebox{0.49 \textwidth}{!}{
        \begin{tabular}{l|cccc|c}
            \toprule
            \textbf{Method} & \textbf{ScienceQA-IMG} & \textbf{POPE} & \textbf{MMBench} & \textbf{MM-Vet} & \textbf{Avg} \\
            \midrule
            Vanilla & 62.5 & 85.7 & 60.2 & 26.9 & 58.8 \\
            \midrule
            + \(1_{1.0} + 3_{1.0}\) & 62.5 & 86.2 & 57.6 & 27.9 & 58.5 \\
            + \(1_{1.0} + 12_{0.25}\) & 62.7 & 86.2 & 57.6 & 27.9 & 58.6 \\
            + \(1_{1.0} + 16_{0.25}\) & 62.8 & 86.2 & 58.7 & 28.1 & 58.9 \\
            + LTDR (\(4_{1.0}\)) & \textbf{63.4} & \textbf{86.6} & \textbf{60.6} & \textbf{29.9} & \textbf{60.1} \\
            \bottomrule
        \end{tabular}
    }
\end{table}


\begin{table}[!t]
    \centering
    \scriptsize
    \caption{Evaluation on high vision-token scenario (NLVR2) using MoE-LLaVA-4Top2 with StableLM-1.6B}
    \label{tab_NLVR2}
    \resizebox{0.46 \textwidth}{!}{
    \begin{tabular}{l|cc}
        \toprule
        \textbf{Method} & \textbf{Accuracy (\%)} & \textbf{Inference Time (s)} \\
        \midrule
        Vanilla & 52.9 & 1121 \\
        \midrule
        Our Method & 54.4 & 1124 \\
        \bottomrule
    \end{tabular}
    }
\end{table}

\subsection{Comparison Studies}
\label{main_comparison_studies}

\textbf{Routing Strategies.}
We compare different routing strategies in Tab.~\ref{tab_routingstrategy}, including task, cluster, instruction, dynamic, conflict mitigation, modality and distribution.
Please refer to the supplementary material for detailed implementations.
Our method achieves state-of-the-art performance, demonstrating its superior performance.

\textbf{Sharpening Vision Token Routing Strategies.} 
We assess the effect of sharpening vision token routing by comparing temperature scaling, entropy regularization, and heuristic approach in Tab.~\ref{tab_sharpening}.
Temperature: Gumbel-Softmax with temperature 0.7.
Entropy: entropy regularization to promote confident routing.
Heuristic: variance constraints on vision token routing distributions.
Simply sharpening vision token routing does not lead to notable gains.
We hypothesize this stems from differing behaviors of head and tail tokens:
Tail tokens, being fewer, benefit from sharper routing to assign them to suitable experts more effectively.
Head tokens, which are abundant and already learn adequately from multiple experts, may adversely affect the learning of tail tokens.

\textbf{Vision Token Selection in VsDEA.}
We choose vision tail tokens (VTTs), whose RPV exceeds the mean RPV of all vision tokens.
Two strategies are compared in the upper of Tab.~\ref{tab_tokenselection}:
Vision head tokens (VHTs), whose RPV are below the mean RPV of all vision tokens.
It selects a large proportion of tokens (\(\approx\)87\%) compared to VTTs (\(\approx\)13\%) .
Instruction-aware Tokens (IATs).
The attention scores between the instruction and vision tokens are used to identify the top 15\% of vision tokens for VsDEA.
As shown in Tab.~\ref{tab_tokenselection}, enhancing expert activations for vision head tokens also improve the model's ability to learn visual information, although the benefits are less significant than those achieved with VTTs.
Moreover, selecting a large proportion of tokens increases the inference time cost.
The influence of IATs is minimal, potentially due to noise from both visual and textual information, which may hinder its ability to reliably identify the important vision tokens.
Finally, we conduct quantitative comparisons against three fixed thresholds in the lower of Tab.~\ref{tab_tokenselection} to further demonstrate the consistent performance advantage of our method.

\textbf{DeepSeekMoE.}
We also compare with DeepSeekMoE~\cite{dai2024deepseekmoe} in Tab.~\ref{tab_deepseek}, which partitions \(K\) experts into \(mK\) experts and activates \(mk\) of them.
Experts are subdivided as \(k_s\) share experts and \(k_r\) routing experts.
Evaluations under \(1_{1.0} + 3_{1.0}\), \(1_{1.0} + 12_{0.25}\) and \(1_{1.0} + 16_{0.25}\) reveal that DeepSeekMoE under performs our method, despite its increase specialization.
This supports our hypothesis that vision and language TER differ, highlighting the need for tailored modality-aware routing.

\subsection{Generalization Studies}

\textbf{Higher Vision-Token Scenario.}
We evaluate inference performance on a multi-image visual question answering task, comparing our method with the baseline in terms of both accuracy and inference speed.
Specifically, we use the NLVR2 dataset~\cite{suhr2019corpus}, in which each sample consists of two images and a textual statement, and the model must determine whether the statement correctly describes both images.
Experimental results, presented in Tab.~\ref{tab_NLVR2}, show that our method outperforms the baseline by 1.5\% in scenarios involving a higher number of vision tokens, while maintaining comparable inference time and introducing no significant latency overhead.

\textbf{Larger Training Dataset.}
We conduct experiments using an expanded training dataset to provide a more comprehensive comparison.
Specifically, we incorporate the Open-LLaVA-NeXT training set, which adds 350K samples to the original 665K samples.
Results summarized in Tab.~\ref{tab_large_instruction} demonstrate that our method achieves greater performance improvements (1.2\%->2.0\%) compared to the baseline after training on larger datasets.
Furthermore, our method on small datasets outperforms the baseline on large datasets.

\begin{table*}[ht]
    \centering
    \scriptsize
    \caption{Validation on large-scale instruction-tuning datasets using MoE-LLaVA-4Top2 with StableLM-1.6B.}
    \label{tab_large_instruction}
    \resizebox{\textwidth}{!}{
        \begin{tabular}{l|c|ccccccc|c}
            \toprule
            \textbf{Method} & \textbf{Data} & \textbf{GQA} & \textbf{ScienceQA-IMG} & \textbf{TextVQA} & \textbf{POPE} & \textbf{MME} & \textbf{MMBench} & \textbf{MM-Vet} & \textbf{Avg} \\
            \midrule
            \multicolumn{10}{c}{\textbf{ShareGPT}~\cite{liu2023llava}} \\
            \midrule
            Vanilla & 665K & 60.3 & 62.6 & 50.1 & 85.7 & 1318.2 & 60.2 & 26.9 & 57.6 \\
            Our Method & 665K & 61.1 & 63.4 & 51.1 & 86.6 & 1363.5 & 60.6 & 29.9 & 58.8 \\
            \midrule
            \multicolumn{10}{c}{\textbf{Open-LLaVA-NeXT}~\cite{chen2024open}} \\
            \midrule
            Vanilla & 1021K & 61.0 & 63.0 & 51.2 & 86.7 & 1360.3 & 61.0 & 29.2 & 58.6 \\
            Our Method & 1021K & 61.4 & 65.4 & 52.5 & 87.3 & 1409.2 & 62.4 & 34.8 & 60.6 \\
            \bottomrule
        \end{tabular}
    }
\end{table*}

\begin{table*}[ht]
    \centering
    \caption{Confidence intervals of the MoE-LLaVA-4Top2 with StableLM-1.6B and Phi-2.7B across seeds.}
    \label{tab_confidence}
    \resizebox{\textwidth}{!}{
        \begin{tabular}{l|ccccccc|c}
            \toprule
            \textbf{Method} & \textbf{GQA} & \textbf{ScienceQA-IMG} & \textbf{TextVQA} & \textbf{POPE} & \textbf{MME} & \textbf{MMBench} & \textbf{MM-Vet} & \textbf{Avg} \\
            \midrule
            StableLM-1.6B & 60.3±0.20 & 62.4±0.20 & 50.0±0.15 & 85.5±0.21 & 1295.1±26.93 & 60.0±0.15 & 26.9±0.40 & 57.5±0.12 \\
            Our Method & 61.1±0.15 & 63.4±0.15 & 51.0±0.10 & 86.6±0.15 & 1342.9±24.66 & 60.7±0.12 & 29.6±0.25 & 58.8±0.06 \\
            \midrule
            Phi2-2.7B & 61.2±0.20 & 68.4±0.21 & 51.3±0.23 & 86.2±0.21 & 1391.9±29.79 & 65.3±0.15 & 34.1±0.15 & 61.1±0.10 \\
            Our Method & 62.2±0.15 & 68.5±0.20 & 52.1±0.10 & 86.7±0.00 & 1414.5±26.45 & 66.7±0.20 & 34.2±0.20 & 61.7±0.06 \\
            \bottomrule
        \end{tabular}
    }
\end{table*}

\subsection{Confidence intervals of the models across different seeds}

We assess StableLM-1.6B and Phi2-2.7B using three random seeds to evaluate the consistency of our method in inference.
As shown in Tab.~\ref{tab_confidence}, our approach consistently outperforms all baselines across these evaluations.

\subsection{Visualization Examples}
\label{app_C_el}

\begin{figure*}[ht]
    \centering
    \includegraphics[width=0.85 \linewidth]{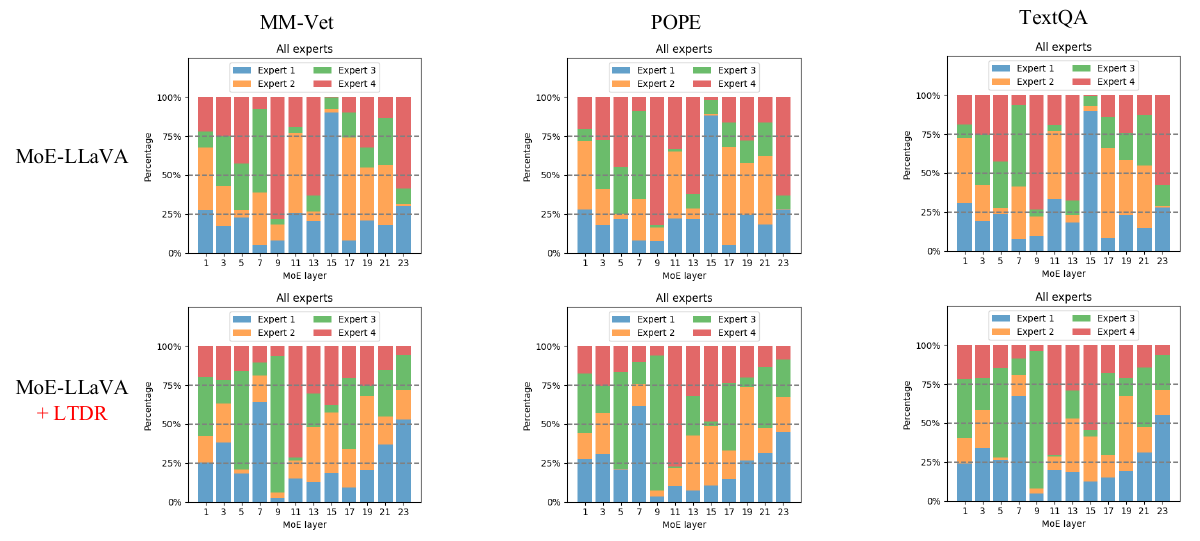}
    \caption{Expert token load across layers. Bar heights indicate token proportions assigned to experts. LTDR yields a more balanced expert utilization.}
    \label{fig_totalload}
\end{figure*}

\begin{figure*}[ht]
    \centering
    \includegraphics[width=0.85 \linewidth]{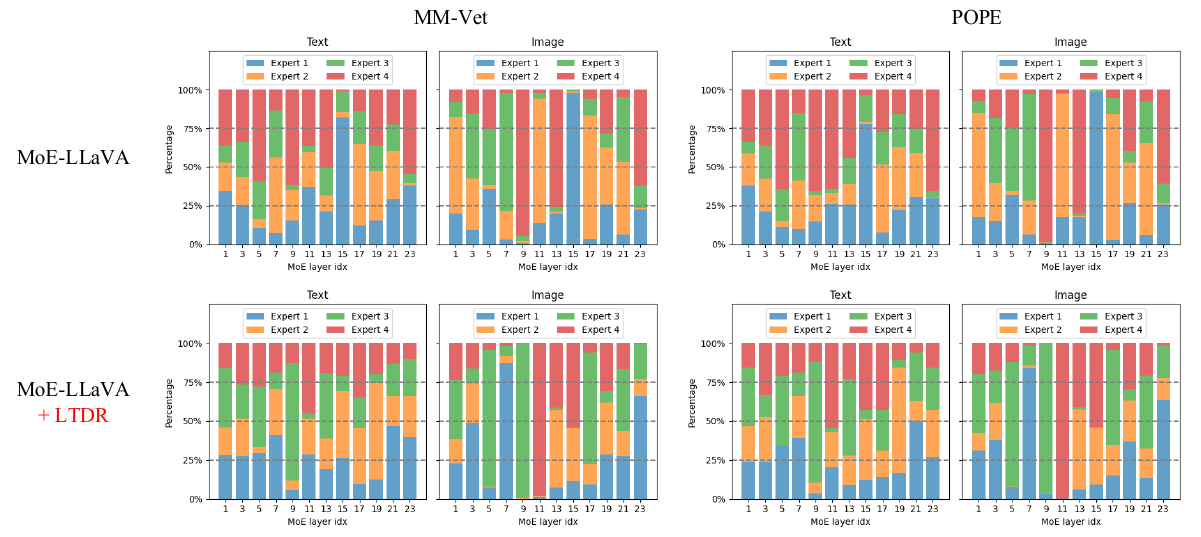}
    \caption{Expert token load cross modal. Bar heights indicate token proportions assigned to experts. LTDR yields more balanced cross-modal expert utilization.}
    \label{fig_modalload}
\end{figure*}

We analyze the distribution of expert load on MoE-LLaVA-4Top2 with StableLM-1.6B.
The expert loads of total tokens and image-text tokens are shown in Fig.~\ref{fig_totalload} $\sim$ Fig.~\ref{fig_modalload}.

Fig.~\ref{fig_totalload} indicates that our method does not significantly amplify the expert load imbalance compared to the baseline, which aligns with the analyses in Fig.~\ref{fig_expertloading}.
Fig.~\ref{fig_modalload} depicts that language tokens follow a relatively uniform distribution.
In contrast, our method eliminates the vision load balancing;
the resulting imbalanced distribution of vision tokens shows that more vision tokens select specialized experts. This supports our hypothesis that the TER for language tokens adheres to a uniform distribution, whereas the TER for vision tokens follows a long-tailed distribution.

\begin{figure*}[!t]
    \centering
    \includegraphics[width=0.85 \linewidth]{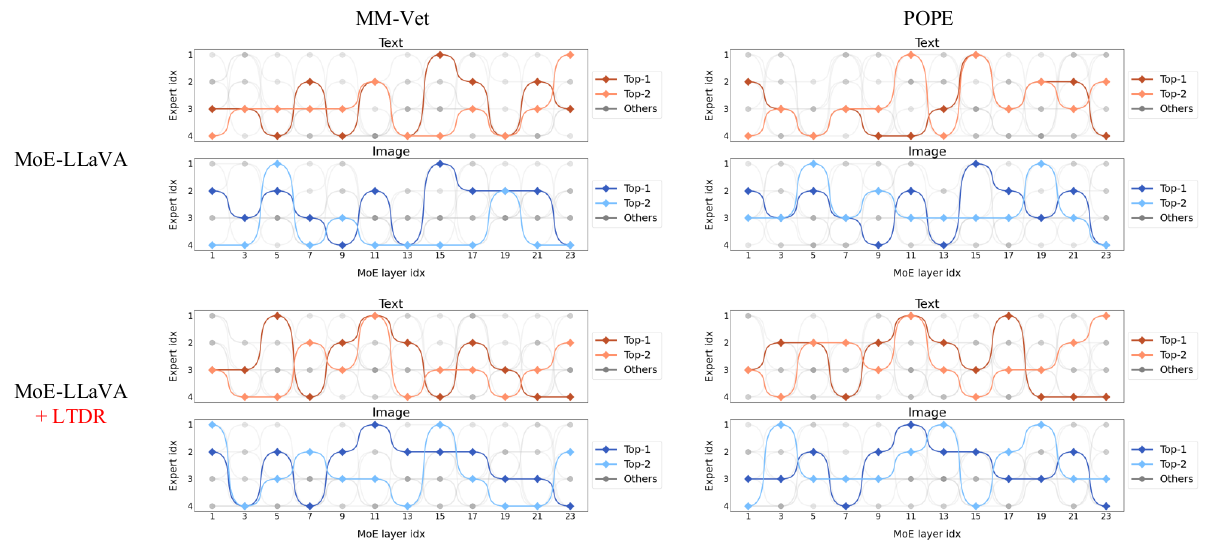}
    \caption{Top-10 pathways for text and image, with Top-2 in color and others in gray. LTDR significantly modifies the original token activation maps.}
    \label{fig_tokenmap}
\end{figure*}

Our method dynamically adjusts token-to-expert paths to improve the language expert load balancing and vision expert specialization, better handling modality-specific distributions.
Fig.~\ref{fig_tokenmap} shows token activation maps. Our method significantly changes the original token activation path.

\section{Conclusion}
\label{main_conclusion}

We reveal the distinct token-to-expert routing (TER) distributions in vision-language tasks: language TER follows a uniform distribution, while vision TER exhibits a long-tailed distribution.
This challenges the traditional load balancing mechanism in MoE:
experts should receive an equal count of tokens to avoid a small number of experts gaining a disproportionately large share of preferences by the router.
To address this, we propose the \textbf{L}ong-\textbf{T}ailed \textbf{D}istribution-aware \textbf{R}outer (LTDR) for vision-language TER, addressing two key challenges:
(1)~Modality-specific distribution-aware routing.
We retain the load balancing mechanism for language TER but abandon it for vision TER, enabling important vision tail tokens to be routed to specialized experts.
(2)~Vision-specific expert activation.
Recognizing the importance of vision tail tokens, we employ a data-augmentation strategy, which increases the number of activated experts to ensure their thorough processing.
To verify the effectiveness of our LTDR, we conduct extensive experiments on both vision-language and vision benchmarks. Experimental results verify the effectiveness of our approach.

\section{Implementation Details}
\label{app_A_implementation}

We utilize the instruction tuning datasets and benchmarks of MoE-LLaVA~\cite{lin2024moellava}, Molmo~\cite{deitke2024molmo} and GMoE~\cite{li2022sparse}.
The composition of these datasets are detailed in Tab.~\ref{tab_details}.
Our training configurations are based on MoE-LLaVA~\cite{lin2024moellava}, Molmo~\cite{deitke2024molmo}, and GMoE~\cite{li2022sparse}. The detailed hyper-parameters and implementation specifics are provided in Tab.~\ref{tab_parameters}.

\begin{table*}[ht]
    \caption{Instruction-tuning datasets and benchmarks on both vision-language and vision tasks.}
    \label{tab_details}
    \centering
    \footnotesize
    \resizebox{1.0 \textwidth}{!}{
    \begin{tabular}{c|c|c|c|c|c}
        \toprule
        \multicolumn{3}{c|}{\textbf{Instruction-tuning Datasets}} & \multicolumn{3}{c}{\textbf{Benchmarks}} \\
        \midrule
        MoE-LLaVA & Molmo & GMoE & MoE-LLaVA & Molmo & GMoE \\
        \midrule
        LLaVA (158K)~\cite{liu2023llava} & VQAv2 (440K)~\cite{goyal2017making} & PACS~\cite{li2017deeper} & GQA~\cite{hudson2019gqa} & ChartQA~\cite{masry2022chartqa} & PACS~\cite{li2017deeper} \\
        ShareGPT (40K)~\cite{sharegpt_chinese_english_90k} & TextVQA (35K)~\cite{singh2019textvqa} & VLCS~\cite{albuquerque2019generalizing} & ScienceQA~\cite{lu2022learn} & DocVQA~\cite{mathew2021docvqa} & VLCS~\cite{albuquerque2019generalizing} \\
        VQAv2 (83K)~\cite{goyal2017making} & OKVQA (9K)~\cite{marino2019ok} & Office-Home~\cite{venkateswara2017deep} & TextVQA~\cite{singh2019textvqa} & AI2D~\cite{kembhavi2016diagram} & Office-Home~\cite{venkateswara2017deep} \\
        GQA (72K)~\cite{hudson2019gqa} & ChartQA (28K)~\cite{masry2022chartqa} & & POPE~\cite{li2023pope} & VQAv2.0~\cite{goyal2017making} & \\
        OKVQA (9K)~\cite{marino2019ok} & DocVQA (39K)~\cite{mathew2021docvqa} & & MME~\cite{fu2023mme} & AndroidControl~\cite{li2406effects} & \\
        OCRVQA (80K)~\cite{mishra2019ocr} & InfographicVQA (24K)~\cite{mathew2022infographicvqa} & & MMBench~\cite{liu2023mmbench} & CountBenchQA~\cite{beyer2024paligemma} & \\
        A-OKVQA (66K)~\cite{schwenk2022okvqa} & AI2D (15K)~\cite{kembhavi2016diagram} & & MM-Vet~\cite{yu2023mmvet} & &  \\
        TextCaps (22K)~\cite{sidorov2019textcaps} & A-OKVQA (17K)~\cite{schwenk2022okvqa} & & & & \\
        RefCOCO (48K)~\cite{kazemzadeh-etal-2014-referitgame} & AndroidControl (300K)~\cite{li2406effects} & & & & \\
        VG (86K)~\cite{krishna2017visual} & ScienceQA (6K)~\cite{lu2022learn} & & & & \\
        & TabWMP (23K)~\cite{lu2022dynamic} & & & & \\
        & ST-VQA (25K)~\cite{biten2019scene} & & & & \\
        & TallyQA (250K)~\cite{acharya2019tallyqa} & & & & \\
        \bottomrule
    \end{tabular}
    }
\end{table*}

\begin{table*}[ht]
    \caption{Detailed training hyper-parameters.}
    \label{tab_parameters}
    \centering
    \resizebox{1.0 \textwidth}{!}{
    \begin{tabular}{ccccc}
        \toprule
        \textbf{Epoch} & \textbf{Learning Rate} & \textbf{Learning Rate Schedule} & \textbf{Weight Decay} & \textbf{Load Balancing Loss Coefficient} \\
        1 & 2e-5 & Cosine & 0.0 & 0.01 \\
        \cline{1-5} \\
        \textbf{Text Max Length} & \textbf{Batch Size per GPU} & \textbf{Train Step} & \textbf{Precision} & \textbf{The Number of \(a \) in VsDEA} \\
        2048 & 16 & original(others) / 2000(Molmo) & Fp16 & 4(others) / 12(Molmo) \\
        \bottomrule
    \end{tabular}
    }
\end{table*}

\section{Additional Comparison Studies}
\label{app_B_comparison}

\subsection{Detail Implementations of Routing Strategies}
\label{app_B_routingstrategy}

\textbf{Task}. Similar to MoLA~\cite{zhou2024exploring}, aims to enhance similar routing for data from the same task while ensure distinct routing for data from different tasks. We conduct experiments using the LLaVA-mix-665k dataset. Empirically, we categorize the data into four task types: Caption, VQA, OCR and Region-aware. Each task type is assigned an expert label: 0 for \emph{Caption}, 1 for \emph{VQA}, 2 for \emph{OCR}, and 3 for \emph{Region-aware}. The dataset distributions are: \emph{Caption} accounts for 3.5\%, \emph{VQA} for 61.6\%, \emph{OCR} for 12.8\%, and \emph{Region-aware} for 22.1\%.

\textbf{Cluster}. Following MoCLE~\cite{gou2023mocle}, we encode instructions from different datasets using the all-MiniLM-L6-v2 \footnote{https://huggingface.co/sentence-transformers/all-MiniLM-L6-v2.} and cluster their embeddings using the \(k\)-means clustering algorithm. After clustering, in line with MoCLE's approach, we initialize \(K\) learnable embeddings, where each embedding corresponds to a cluster center. When a sample belongs to the \(k\)-th cluster center, the \(k\)-th learnable embedding is extracted and passed to the router to predict routing scores. We set \(k\)-th=128, consistent with MoCLE's practice, and do not incorporate the load balancing loss.

\textbf{Instruct}. Following the approach of LoRA-MoE~\cite{chen2311octavius}, we compute the average of instruction token representations for each instance and use this as input to predict its routing scores across experts. Based on these routing scores, the Top-\(k\) experts are selected for each sample to generate the final prediction. In alignment with LoRA-MoE's methodology, we do not include the load balancing loss in our implementation.

\textbf{Dynamic}. Drawing inspiration from DYNMOE~\cite{guo2024dynamic}, we highlight its gating mechanism, which allows tokens to dynamically determine the number of experts to activate. Additionally, an adaptive process automatically adjusts the number of experts during training. We reference its experimental results on the MoE-LLaVA-4Top2 with StableLM-1.6B model, where an average of 1.25 experts out of 4 are activated per token.

\textbf{STGC}. STGC~\cite{yang2024stgc} employs token-level gradients to identify conflicting tokens within experts. Additionally, it introduces a regularization loss designed to encourage conflicting tokens to route away from their current experts to alternative ones, thereby minimizing interference among tokens within the same expert. We reference their experimental results on the MoE-LLaVA-4Top2 using the StableLM-1.6B model.

\textbf{Modality}. Following MoMa~\cite{lin2024moma}, we partition experts into two groups dedicated to vision and language, respectively. To maintain the total number of experts \(K\) (set to 4) and the number of activated experts \(k\) (set to 2), we designate 2 experts as vision experts and the remaining 2 as language experts, activating 1 expert from each modality group.

\textbf{Distribution}. Following RIDE~\cite{wang2020long}, we cluster tokens into six categories via k-means (matching the 4Top2 expert paths, \(C^2_4\). Each category is mapped to a specific expert pair (e.g., category 1 → experts 1 \& 2). Tokens of each category are directed to their assigned expert pair.

\begin{figure}[ht]
    \centering
    \includegraphics[width=1.0\linewidth]{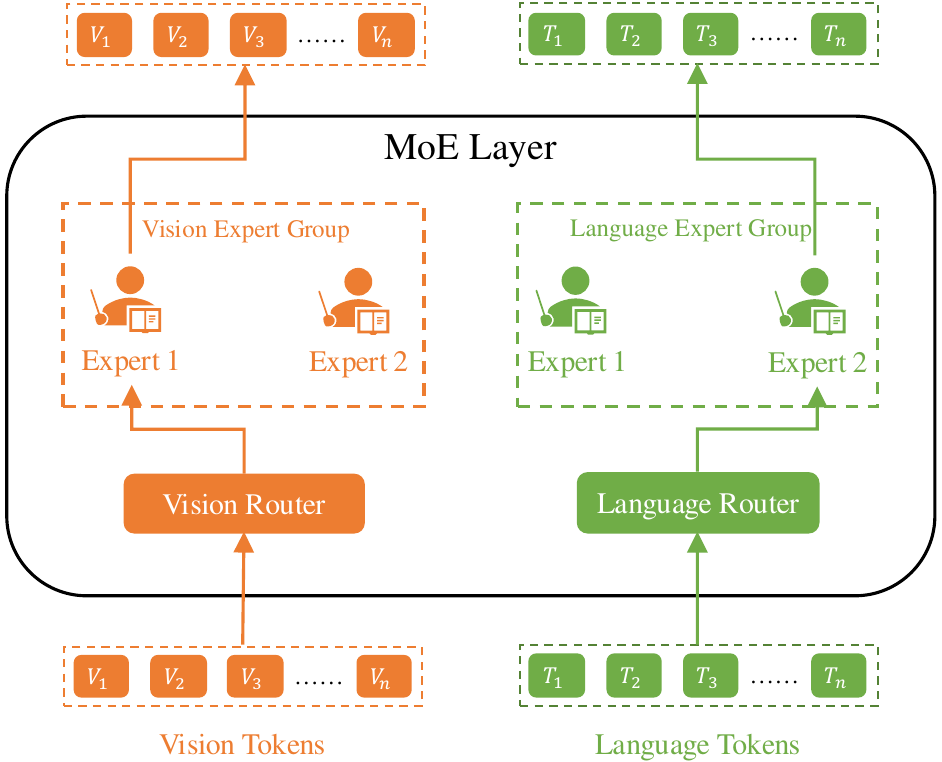}
    \caption{The modality-aware MoE architecture which divides experts for vision and language.}
    \label{fig_modalityaware}
\end{figure}

\subsection{Modality-aware MoE Architecture}
\label{app_B_modalityaware}


To compare with modality-aware MoE architectures~\cite{chen2024eve, lin2024moma, nguyen2024expert} as shown in Fig.~\ref{fig_modalityaware}, we adopt the following configurations:

\textbf{MoE-LLaVA-v2Top1-t2Top1.} We partition experts into two groups dedicated to vision and language. As illustrated in Fig.~\ref{fig_modalityaware}, to maintain the total number of experts \(K\) (set to 4) and the number of activated experts \(k\) (set to 2), we designate 2 experts as vision experts and the remaining 2 as language experts, activating 1 expert from each modality group.

\textbf{MoE-LLaVA-v4Top2-t4Top2.} We expand the 4 experts into 8 by splitting FFN intermediate hidden dimension~\cite{dai2024deepseekmoe}, assigning 4 as vision experts and the remaining 4 as language experts, and activating 2 experts from each modality group.

\textbf{MoE-LLaVA-v2Top1-t2Top1-MsDaR.} Similar to \emph{MoE-LLaVA-v2Top1-t2Top1}, but it removes the expert load balancing constraint for the vision expert group (similar to the distribution-aware router in our proposed approach, LTDR).

\textbf{MoE-LLaVA-v2Top1-t2Top1-MsDaR-shared.} Akin to the former, but includes a shared expert for world knowledge~\cite{dai2024deepseekmoe}.

The results in Tab.~\ref{tab_comparison_modalityaware} indicate that the modality-aware MoE does not enhance the performance of MoE-LLaVA.
Moreover, increasing the number of experts results in a performance decline, suggesting that a larger number of experts exacerbates expert load balancing issues, which negatively impacts vision TER.
When the expert load balancing constraint is removed for the vision expert group, \emph{MoE-LLaVA-v2Top1-t2Top1-MsDaR} shows improved performance compared to MoE-LLaVA-4Top2, validating the effectiveness of our MsDaR module.
Finally, the addition of an extra expert in \emph{MoE-LLaVA-v2Top1-t2Top1-MsDaR-shared} does not yield benefits, highlighting the distinctions between vision-language MoE and language MoE~\cite{dai2024deepseekmoe}.

\begin{table*}[ht]
    \caption{Comparison with modality-aware MoE~\cite{chen2024eve, lin2024moma, nguyen2024expert} on MoE-LLaVA-4Top2 with StableLM-1.6B.}
    \label{tab_comparison_modalityaware}
    \centering
    \resizebox{1.0 \textwidth}{!}{
    \begin{tabular}{l|ccccccc|c}
        \toprule
        \textbf{Method} & \textbf{GQA} & \textbf{ScienceQA-IMG} & \textbf{TextVQA} & \textbf{POPE} & \textbf{MME} & \textbf{MMBench} & \textbf{MM-Vet} & \textbf{Avg} \\
        \midrule
        MoE-LLaVA-4Top2 & 60.3 & 62.6 & 50.1 & 85.7 & 1318.2 & 60.2 & 26.9 & 57.6 \\
        \midrule
        MoE-LLaVA-v2Top1-t2Top1 & 60.4 & 61.6 & 49.2 & 85.9 & 1293.3 & 61.1 & 28.4 & 57.7 \\
        MoE-LLaVA-v4Top2-t4Top2 & 60.3 & 58.6 & 46.8 & 85.7 & 1296.6 & 55.4 & 26.4 & 55.5 \\
        MoE-LLaVA-v2Top1-t2Top1-MsDaR & 60.9 & 61.3 & 51.0 & 86.5 & 1324.5 & 61.1 & 28.4 & 58.2 \\
        MoE-LLaVA-v2Top1-t2Top1-MsDaR-shared & 61.0 & 62.5 & 51.3 & 86.5 & 1333.8 & 60.0 & 28.0 & 58.2 \\
        \midrule
        Our Method & 61.1 & 63.4 & 51.1 & 86.6 & 1363.5 & 60.6 & 29.9 & 58.8 \\
        \bottomrule
    \end{tabular}
    }
\end{table*}

\section{Additional Corroborating Studies}
\label{app_F_response}

\subsection{Strategy-swap Ablation on Vision and Language}

We perform a strategy-swap ablation on the language-side while keeping the vision-side fixed.
As shown in Tab.~\ref{tab_strategyswap}, removing the load balancing loss on the language side (language+MsDaR) results in fluctuating performance, yielding a 
little improvement over the vanilla model.
Moreover, introducing text-specific dynamic expert activation (TsDEA) does not lead to consistent performance gains.
In contrast, removing the load balancing loss on the vision side (vision+MsDaR) yields stable improvements, and further incorporating vision-specific dynamic expert activation (VsDEA) provides additional gains, achieving an average improvement of 1.2\%.

\begin{table*}[ht]
    \centering
    \scriptsize
    \caption{Strategy-swap ablation studies on vision and language modalities.}
    \label{tab_strategyswap}
    \resizebox{1.0 \textwidth}{!}{
    \begin{tabular}{l|ccccccc|c}
        \toprule
        \textbf{MoE-LLaVA\_StableLM} & \textbf{GQA} & \textbf{ScienceQA-IMG} & \textbf{TextVQA} & \textbf{POPE} & \textbf{MME} & \textbf{MMBench} & \textbf{MM-Vet} & \textbf{Avg} \\
        \midrule
        Vanilla & 60.3 & 62.6 & 50.1 & 85.7 & 1318.2 & 60.2 & 26.9 & 57.6 \\
        \midrule
        \multicolumn{9}{c}{\textbf{Language}} \\
        \midrule
        + MsDaR & 60.8 & 62.0 & 50.2 & 85.9 & 1254.1 & 60.0 & 28.0 & 57.8 \\
        \midrule
        + MsDaR\&TsDEA & 60.1 & 61.2 & 50.4 & 86.2 & 1282.4 & 60.1 & 28.6 & 57.8 \\
        \midrule
        \multicolumn{9}{c}{\textbf{Vision}} \\
        \midrule
        + MsDaR & 61.1 & 62.3 & 51.2 & 86.6 & 1324.3 & 59.9 & 27.9 & 58.2 \\
        \midrule
        + MsDaR\&VsDEA & 61.1 & 63.4 & 51.1 & 86.6 & 1363.5 & 60.6 & 29.9 & 58.8 \\
        \bottomrule
    \end{tabular}
    }
\end{table*}

\subsection{Broader Cross-Router Performance}

In addition to comparing the different backbones in Section.IV-B, we also report the cross-router results, as summarized in Tab.~\ref{tab_cross_router}.
In the Top-Dynamic setting, expert activation is determined by token-score ranking and expert capacity, where the expert capacity is defined as $num_{\text{token}} / num_{\text{experts}} \times 3$.
LTDR with Top-1 achieves a 0.9\% improvement over the vanilla model. Although dynamic routers provide additional gains, their performance remains lower than that of LTDR combined with Top-a.

\begin{table*}[ht]
    \centering
    \scriptsize
    \caption{Broader cross-router performance across different expert activation numbers.}
    \label{tab_cross_router}
    \resizebox{1.0 \textwidth}{!}{
    \begin{tabular}{l|ccccccc|c}
        \toprule
        \textbf{MoE-LLaVA\_StableLM} & \textbf{GQA} & \textbf{ScienceQA-IMG} & \textbf{TextVQA} & \textbf{POPE} & \textbf{MME} & \textbf{MMBench} & \textbf{MM-Vet} & \textbf{Avg} \\
        \midrule
        Vanilla+Top-1 (k=1)	& 58.6 & 55.8 & 45.0 & 85.2 & 1245.3 & 56.2 & 27.2 & 54.7 \\
        \midrule
        Vanilla+Top-2 (k=2) & 60.3 & 62.6 & 50.1 & 85.7 & 1318.2 & 60.2 & 26.9 & 57.6 \\
        \midrule
        LTDR+Top-1 (k=1) & 59.7 & 58.2 & 45.6 & 85.8 & 1302.9 & 57.1 & 27.0 & 55.6 \\
        \midrule
        LTDR+Dynamic (avg k=3) & 60.8 & 62.1 & 51.1 & 86.8 & 1332.8 & 60.2 & 28.9 & 58.3 \\
        \midrule
        LTDR+Top-a (avg k=2.26) & 61.1 & 63.4 & 51.1 & 86.6 & 1363.5 & 60.6 & 29.9 & 58.8 \\
        \bottomrule
    \end{tabular}
    }
\end{table*}

\subsection{Impact of Reducing Load Balancing}

We analyze the effect of reducing load balancing, with the results presented in Tab.~\ref{tab_reducing}.
Specifically, we decrease the vision-side load balancing coefficient from 0.01 to 0.001.
The results indicate that this adjustment is less effective than removing load balancing entirely.

\begin{table*}[ht]
    \centering
    \scriptsize
    \caption{Impact of reducing load balancing.}
    \label{tab_reducing}
    \resizebox{1.0 \textwidth}{!}{
    \begin{tabular}{l|ccccccc|c}
        \toprule
        \textbf{MoE-LLaVA\_StableLM} & \textbf{GQA} & \textbf{ScienceQA-IMG} & \textbf{TextVQA} & \textbf{POPE} & \textbf{MME} & \textbf{MMBench} & \textbf{MM-Vet} & \textbf{Avg} \\
        \midrule
        Vanilla & 60.3 & 62.6 & 50.1 & 85.7 & 1318.2 & 60.2 & 26.9 & 57.6 \\
        \midrule
        Reducing-based & 60.2 & 62.4 & 50.9 & 86.3 & 1315.7 & 59.1 & 28.4 & 57.8 \\
        \midrule
        LTDR & 61.1 & 63.4 & 51.1 & 86.6 & 1363.5 & 60.6 & 29.9 & 58.8 \\
        \bottomrule
    \end{tabular}
    }
\end{table*}

\begin{table*}[ht]
    \centering
    \scriptsize
    \caption{Generalizability across different models.}
    \label{tab_generalizability}
    \resizebox{1.0 \textwidth}{!}{
    \begin{tabular}{l|ccccccc|c}
        \toprule
        \textbf{MoE-LLaVA\_StableLM} & \textbf{GQA} & \textbf{ScienceQA-IMG} & \textbf{TextVQA} & \textbf{POPE} & \textbf{MME} & \textbf{MMBench} & \textbf{MM-Vet} & \textbf{Avg} \\
        \midrule
        Phi-2B with 10\% & 62.1 & 68.3 & 51.6 & 86.5 & 1406.3 & 65.6 & 33.7 & 61.3 \\
        \midrule
        Phi-2B with 15\% & 62.0 & 68.4 & 51.8 & 86.6 & 1423.1 & 66.3 & 34.0 & 61.5 \\
        \midrule
        Phi-2B with 20\% & 61.3 & 67.8 & 51.3 & 86.2 & 1414.5 & 66.5 & 34.1 & 61.2 \\
        \midrule
        Phi-2B with mean-RPV & 62.2 & 69.3 & 52.9 & 87.5 & 1446.5 & 66.8 & 34.9 & 62.3 \\
        \bottomrule
    \end{tabular}
    }
\end{table*}

\subsection{Generalizability across datasets/models}

We compare the mean RPV using thresholds of 10\%, 15\%, and 20\%, as the observed mean RPV (13\%) falls within this range.
This allows us to visually assess how small variations in the threshold affect performance.
We evaluate these fixed-proportion thresholds on the MoE-LLaVA with StableLM-1.6B model across multiple test datasets, as detailed in Appendix~\ref{app_B_comparison}.
Here, we further present generalizability experiments on the MoE-LLaVA with Phi-2-2.7B model, as shown in Tab.~\ref{tab_generalizability}.
The results show that mean-RPV consistently outperforms the comparison thresholds.
We also compare mean-RPV with the learnable strategy in Tab.~\ref{tab_tuning};
however, the tuning method yields performance that lies between that of the original model and the LTDR-based model.

\begin{table*}[ht]
    \centering
    \scriptsize
    \caption{Comparison with systematic tuning policy.}
    \label{tab_tuning}
    \resizebox{1.0 \textwidth}{!}{
    \begin{tabular}{l|ccccccc|c}
        \toprule
        \textbf{MoE-LLaVA\_StableLM} & \textbf{GQA} & \textbf{ScienceQA-IMG} & \textbf{TextVQA} & \textbf{POPE} & \textbf{MME} & \textbf{MMBench} & \textbf{MM-Vet} & \textbf{Avg} \\
        \midrule
        Vanilla & 60.3 & 62.6 & 50.1 & 85.7 & 1318.2 & 60.2 & 26.9 & 57.6 \\
        \midrule
        Tuning-based & 61.0 & 62.1 & 50.6 & 86.3 & 1322.3 & 59.9 & 28.9 & 58.1 \\
        \midrule
        LTDR & 61.1 & 63.4 & 51.1 & 86.6 & 1363.5 & 60.6 & 29.9 & 58.8 \\
        \bottomrule
    \end{tabular}
    }
\end{table*}

\subsection{The link between RPV and token informativeness}


\begin{table}[ht]
    \centering
    \scriptsize
    \caption{Statistics-based analysis.}
    \label{tab_statistics}
    \resizebox{0.49 \textwidth}{!}{
    \begin{tabular}{l|c}
        \toprule
        \textbf{MoE-LLaVA\_StableLM} & \textbf{mean L2 norm} \\
        \midrule
        Top-13\% & 0.3158 \\
        \midrule
        Top-13\% - Top-26\% & 0.2124 \\
        \midrule
        Top-26\% - Top-39\% & 0.1475 \\
        \bottomrule
    \end{tabular}
    }
\end{table}

We evaluate the link between RPV and token informativeness from two aspects:
1) Performance-based analysis (interpret ability).
Higher model performance suggests that the corresponding tokens carry richer information.
As described in Section.IV-D, we sort tokens by RPV and classify those above the mean RPV as vision tail tokens (13\%) and the remaining ones as vision head tokens (87\%).
Although vision head tokens are far more numerous than tail tokens, applying the VsDEA strategy to head tokens yields substantially lower performance than applying it to tail tokens.
This performance gap indicates that high-RPV vision tail tokens encode more informative content.
2) Statistics-based analysis (statistical).
A token group with a higher mean L2 norm of its vector representations reflects richer underlying information.
We compare the mean L2 norm of the top 13\% vision token vectors (vision tail tokens) with those of tokens ranked between the top 13\%–26\% and 26\%–39\% by RPV.
As shown in Tab.~\ref{tab_statistics}, top 13\% exhibits a higher mean L2 norm than the latter two, providing additional evidence that high-RPV vision tail tokens carry richer information.

\section{Visualization Examples}
\label{app_C_visualization}

\subsection{Routing Probability Variance}
\label{app_C_rpv}

\begin{figure*}[ht]
    \centering
    \includegraphics[width=0.90 \linewidth]{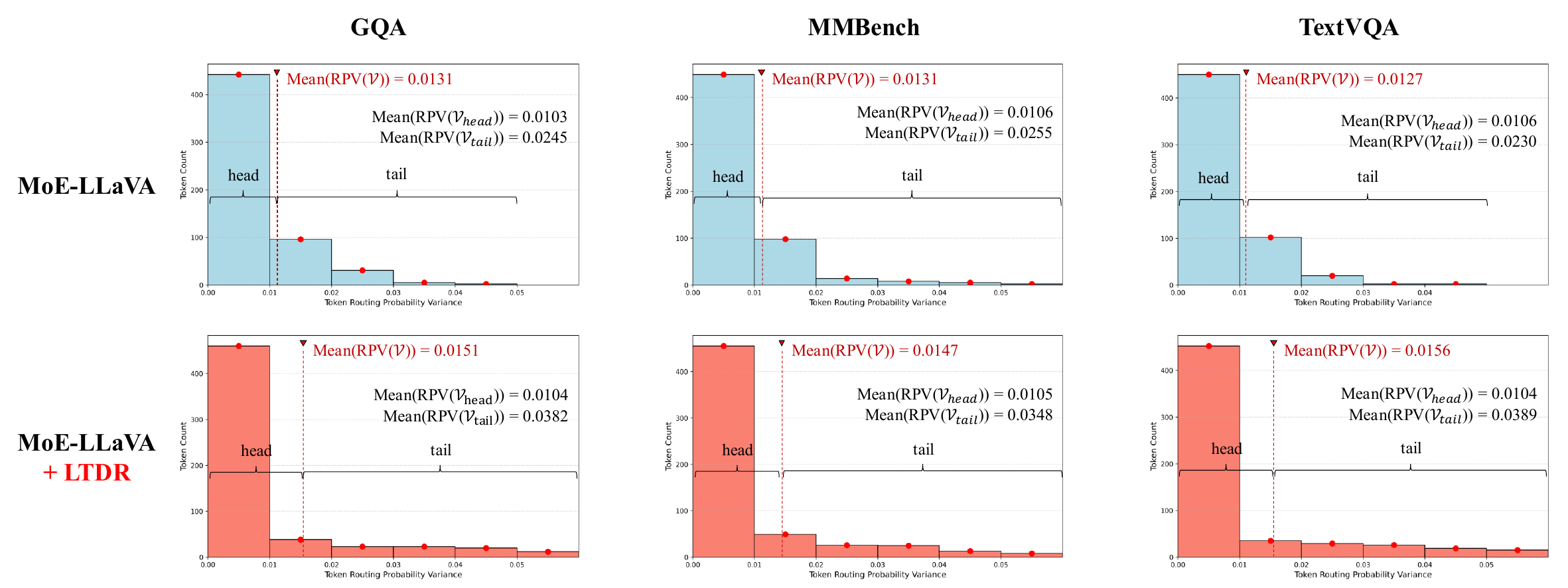}
    \caption{The Routing probability variance distribution of vision head tokens and vision tail tokens.}
    \label{fig_distribution}
\end{figure*}

\begin{figure*}[ht]
    \centering
    \includegraphics[width=0.90 \linewidth]{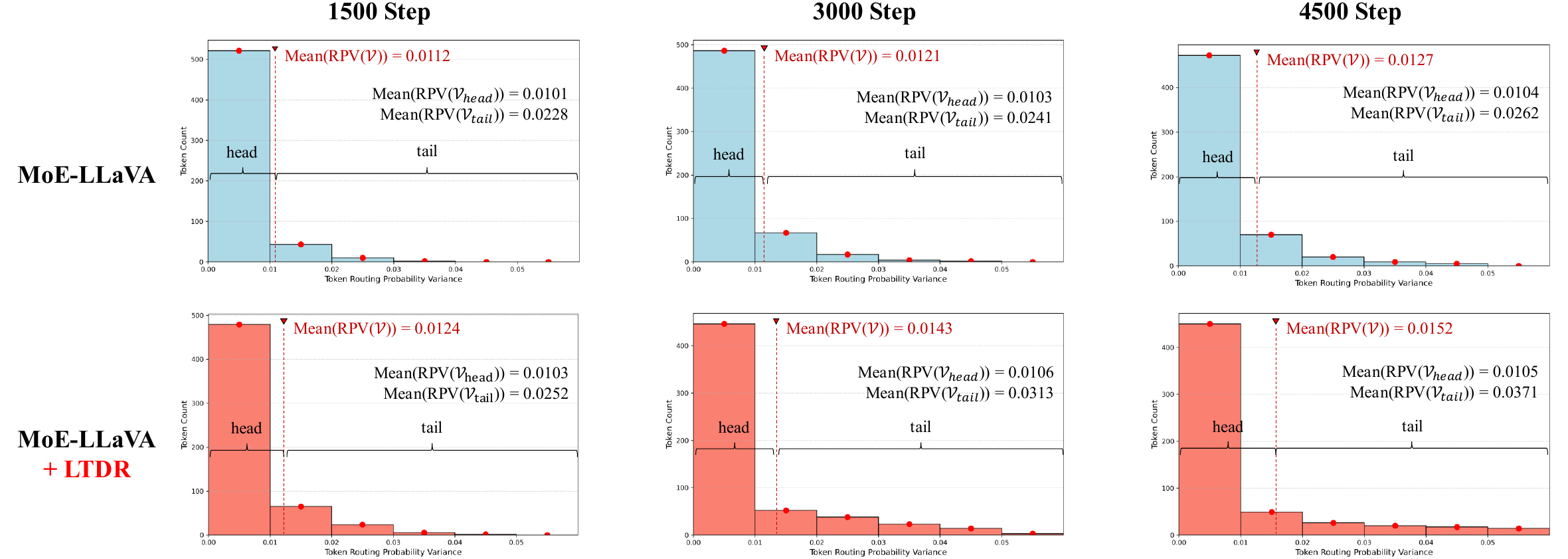}
    \caption{The Routing probability variance distribution at three training steps on TextVQA.}
    \label{fig_distribution_train}
\end{figure*}

\begin{figure*}[ht]
    \centering
    \includegraphics[width=0.90 \linewidth]{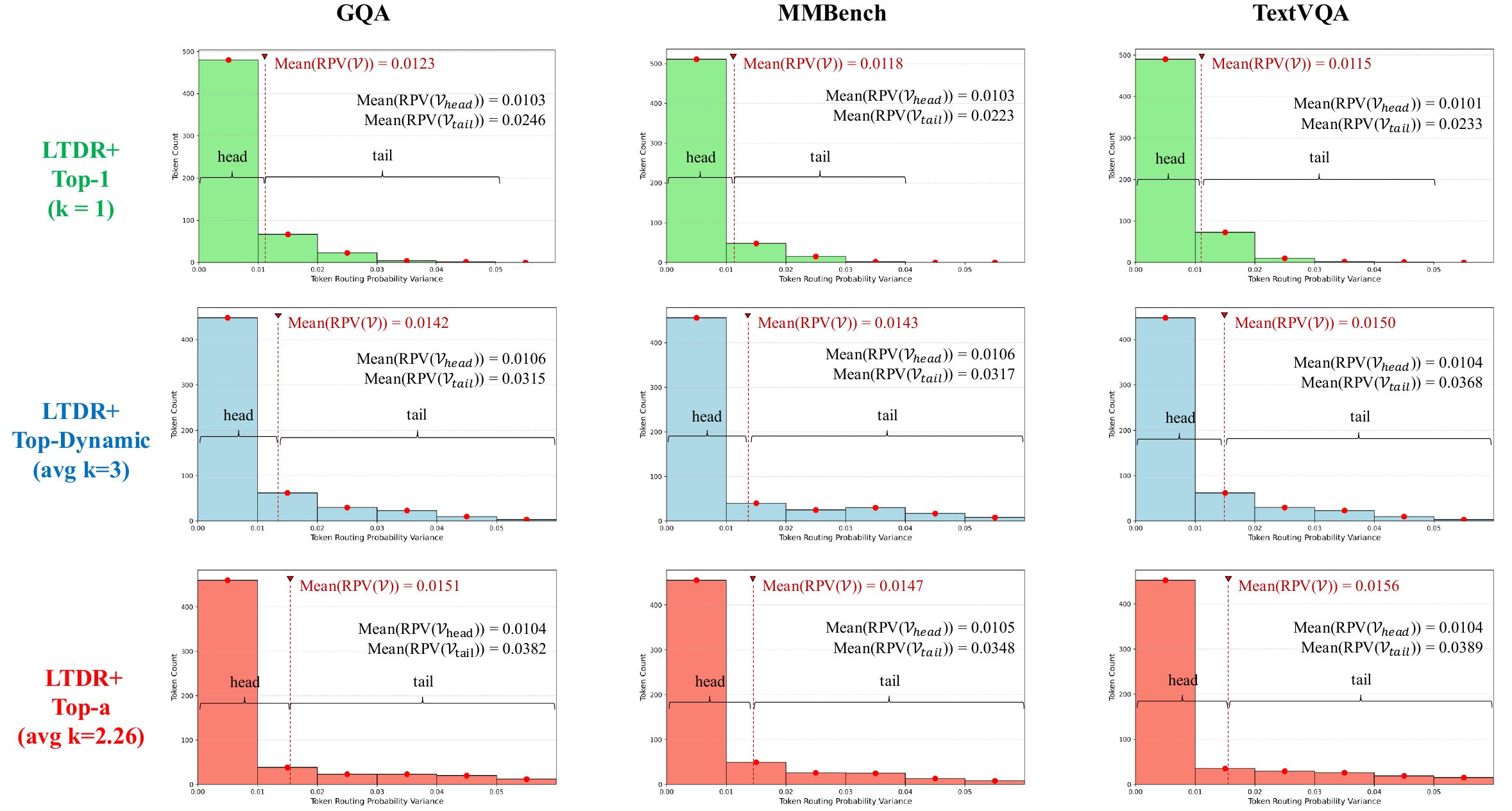}
    \caption{The Routing probability variance distribution cross routers.}
    \label{fig_distribution_router}
\end{figure*}

We also compare the token routing probability variance (RPV) between vision head tokens and vision tail tokens across GQA~\cite{hudson2019gqa}, MMBench~\cite{liu2023mmbench} and TextVQA~\cite{singh2019textvqa}.
As shown in Fig.~\ref{fig_distribution}, for images that yield 576 tokens from CLIP, \(\mathrm{Mean}(\mathrm{RPV}(\mathcal{V})) \) denotes the mean RPV of all vision tokens \(\mathcal{V} \), \(\mathcal{V}_{head}\) and \(\mathcal{V}_{tail}\) denote vision head tokens and vision tail tokens.
The bars in the figure is the mean token count with RPV ranging from left to right for images (\emph{e.g.}, in the upper left figure, the count of tokens with RPV ranging from 0.00 to 0.01 is 442).
Our method significantly increases the mean RPV of vision tail tokens.
Given that RPV reflects the TER probability distribution, these results demonstrate that vision tail tokens gain the ability to select their specialized experts.
Meanwhile, the mean RPV of vision head tokens remains unchanged, indicating that vision head tokens are not affected.
Moreover, we add three sets of RPV distribution visualization comparisons: training stages RPV (Fig.~\ref{fig_distribution_train}), and cross-router RPV (Fig.~\ref{fig_distribution_router}).
The RPV distributions at 1,500, 3,000, and 4,500 training steps on TextVQA indicate that, although both the baseline and our method improve mean RPV, our approach achieves substantially greater enhancement.
This suggests that the baseline has limited capacity to effectively capture critical visual information.
Cross-Router:
In the Top-Dynamic setting, expert activation is determined by token score ranking and expert capacity, where expert capacity is defined as $num_{\text{token}} / num_{\text{experts}} \times 3$.
The results indicate that the mean RPV of the vision tail under Top-1 is significantly lower than in other configurations, suggesting that increasing the number of parameters can enhance vision tail token learning. 
However, despite Top-Dynamic having the largest number of parameters, it does not outperform Top-a in terms of vision tail RPV, implying that simply increasing parameter count does not guarantee performance improvement.

\subsection{Visualization Cases.}
\label{app_C_vc}

\begin{figure*}[ht]
    \centering
    \includegraphics[width=0.8\linewidth]{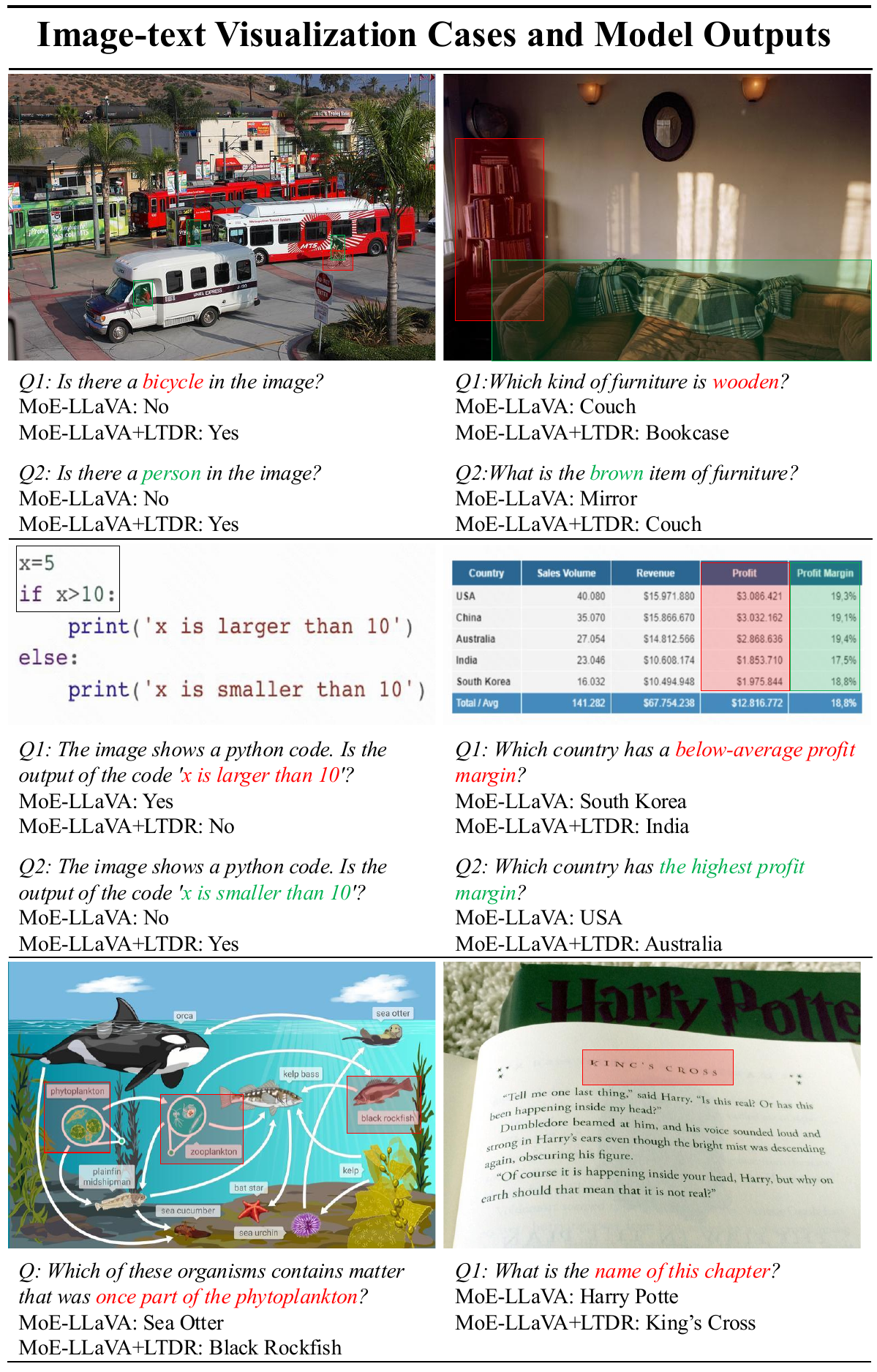}
    \caption{Visualization Cases on benchmarks of MoE-LLaVA.}
    \label{fig_vc}
\end{figure*}

Our visualizations in Fig.~\ref{fig_vc} reveal that vision tail tokens focus on critical instruction-aware image patches, capturing question-relevant visual information which enhances answer accuracy.

\newpage
\clearpage
\balance
\bibliographystyle{IEEEtran}
\bibliography{main}

\end{document}